\documentclass[10pt,twocolumn,letterpaper]{article}

\usepackage[applications, pagenumbers]{wacv}      

\usepackage[accsupp]{axessibility}
\usepackage{amsmath}
\usepackage{subcaption}
\DeclareMathOperator*{\argmin}{arg\,min}
\newcommand{\supp}{\mathop{\mathtt{supp}}}
\newcommand{\reals}{\ensuremath\mathbb{R}}
\usepackage{amssymb}
\usepackage{mathtools}
\usepackage{xcolor}
\usepackage{nicefrac}
\usepackage{algorithm}
\usepackage{algorithmic}
\newcommand{\Assign}{\ensuremath{\leftarrow}}

\definecolor{wacvblue}{rgb}{0.21,0.49,0.74}
\usepackage[pagebackref,breaklinks,colorlinks,allcolors=wacvblue]{hyperref}

\def\wacvPaperID{754} 
\def\confName{WACV}
\def\confYear{2026}

\title{SSplain: Sparse and Smooth Explainer for Retinopathy of Prematurity Classification}

\author{
Elifnur Sunger\textsuperscript{1}, 
Tales Imbiriba\textsuperscript{2},
Peter Campbell\textsuperscript{3}, 
Deniz Erdogmus\textsuperscript{1},
Stratis Ioannidis\textsuperscript{1}, 
Jennifer Dy\textsuperscript{1} 
\\[1ex]
\textsuperscript{1}Northeastern University, Boston, MA, USA\\
\textsuperscript{2}University of Massachusetts Boston, Boston, MA, USA\\
\textsuperscript{3}Oregon Health \& Science University, Portland, OR, USA\\
{\tt\small \textsuperscript{1}\{sunger.e, d.erdogmus, e.ioannidis, j.dy\}@northeastern.edu,  \textsuperscript{2}tales.imbiriba@umb.edu,} \\{\tt\small\textsuperscript{3}campbelp@ohsu.edu}
}
\begin{document}
\maketitle
\begin{abstract}
Neural networks are frequently used in medical diagnosis. However, due to their black-box nature, model explainers are used to help clinicians understand better and trust model outputs. This paper introduces an explainer method for classifying Retinopathy of Prematurity (ROP) from fundus images. Previous methods fail to generate explanations that preserve input image structures such as smoothness and sparsity. We introduce \textit{Sparse and Smooth Explainer (SSplain)}, a method that generates pixel-wise explanations while preserving image structures by enforcing smoothness and sparsity. This results in realistic explanations to enhance the understanding of the given black-box model. To achieve this goal, we define an optimization problem with combinatorial constraints and solve it using the Alternating Direction Method of Multipliers (ADMM). Experimental results show that SSplain outperforms commonly used explainers in terms of both post-hoc accuracy and smoothness analyses. Additionally, SSplain identifies features that are consistent with domain-understandable features that clinicians consider as discriminative factors for ROP. We also show SSplain's generalization by applying it to additional publicly available datasets. Code is available at \url{https://github.com/neu-spiral/SSplain}.
\end{abstract}

\section{Introduction}
\label{sec:introduction}
Retinopathy of Prematurity (ROP) affects premature infants and is a leading cause of childhood blindness \cite{gilbert2001childhood}. Early detection and treatment can prevent vision impairment; this has motivated the study of automated, machine-learning based methods for detecting the disease. 
Early work on ROP detection from fundus images \cite{ataer2015computer,tian2016toward} used signal processing techniques to extract clinically relevant features; \textit{dilation}, i.e., thickness of vessels in the retina, and \textit{tortuosity}, as captured by vessel curvature. Generally, the more dilated and tortuous the vessels, the more severe the ROP~\cite{chiang2021international}. 

While these approaches lead to interpretable algorithms, recent research has improved classification performance by employing convolutional neural network (CNN) architectures~\cite{brown2018automated,worrall2016automated} combined with attention-based mechanisms~\cite{yildiz2021structurally, yildiz2021structural}. For example, Brown \etal ~\cite{brown2018automated} propose a state-of-the-art ROP detection CNN based on the Inception architecture~\cite{szegedy2015going} that achieves $91\%$ balanced accuracy. However, due to the black-box nature of neural networks, clinicians often hesitate to trust the diagnostic outcomes of such deep learning methods~\cite{hanif2021applications}. 

As an attempt to explain classifiers' decisions, the so-called black-box explainers~\cite{fong2017interpretable,ribeiro2016should,simonyan2013deep} have been frequently used. When applied to image classifiers, they produce \emph{explanation maps}, i.e., scores of pixels indicating their relevance to prediction.

\begin{figure*}[t!]
    \centering 
    \includegraphics[width=1.0\linewidth]{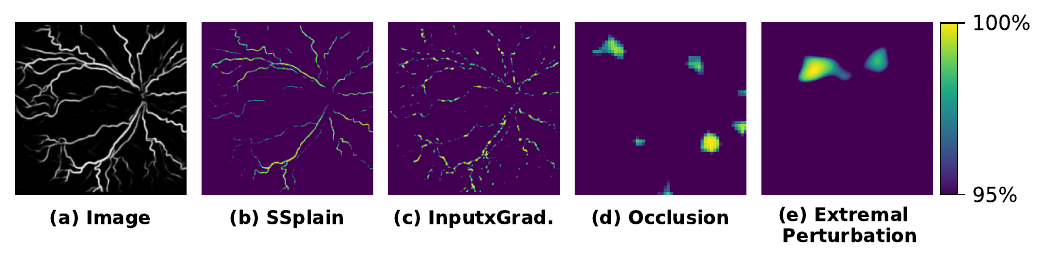}
    \caption{Explanation maps of (a) a sample image with ROP using (b) SSplain, (c) Input$\times$Gradient~\cite{shrikumar2016not}, (d) Occlusion~\cite{zeiler2014visualizing} and (e) Extremal Perturbation~\cite{fong2019understanding} methods. 
    We visualize only the top 5\% of pixels, as ranked by the corresponding explanation maps, coloring the remaining pixels as purple.
    SSplain generates sparse and contiguous regions of importance, that are also more dilated and tortuous than obscured vessels.  Sparse method Input$\times$Gradient focuses on similar areas but it is highly fragmented. Occlusion and Extremal Perturbation select discontinuous regions: tortuous vessels and dilated vessels, respectively. Both, however, fail to preserve vessel structure.}
    \label{fig:sample figure}
\end{figure*}

Unfortunately, generic explainers fail to incorporate domain-specific image structures that might aid in interpretability. For example, fundus images, as shown in Figure~\ref{fig:sample figure}(a), exhibit \textit{sparsity}: images consist mostly of black pixels corresponding to the background, and discriminative features are concentrated on a subset of the vessels. Another relevant property is \textit{smoothness}: regions of importance form contiguous segments of proximal vessel pixels of similarly (elevated) dilation and tortuosity. Black-box explainers for ROP detection need to consider and preserve the input image structure, such as smoothness and sparsity, to generate realistic explanation maps or attribution scores.  

We introduce a Sparse and Smooth Explainer (SSplain) for ROP, which generates pixel-wise sparse and smooth explanations for the given black-box model and exploits the inherent image structure. Our contributions are as follows:
\begin{itemize}
    \item We propose an explainability method that preserves inherent ROP image structures by enforcing explainer smoothness and sparsity. This approach enhances the understanding of the black-box predictor for classifying ROP images by highlighting relevant structures. 
    \item This problem is non-convex and has combinatorial constraints; we use the Alternating Direction Method of Multipliers (ADMM)~\cite{boyd2011distributed} to reduce this optimization to tractable subproblems.
    \item Through post-hoc performance and smoothness analyses, we demonstrate that SSplain outperforms several state-of-the-art competitors.
    \item We evaluate the explainers on two domain-specific metrics, dilation and tortuosity, which are discriminative factors for ROP. SSplain \emph{consistently identifies more dilated and tortuous regions than competitors}, accentuating the superiority of SSplain towards ROP explainability. We stress that this occurs even though the tortuosity and dilation are neither explicitly measured nor incorporated in SSplain design objective. 
    \item We perform a sanity check test~\cite{adebayo2018sanity} for explainability methods, showing that SSplain is sensitive to changes in the model, as desired.
    \item We apply SSplain to MNIST~\cite{lecun1998mnist} and FMNIST~\cite{xiao2017fashion}, demonstrating its generalization and outperforming state-of-the-art methods in post-hoc analyses.
    
\end{itemize}

\section{Related work}
Numerous methods have been proposed to explain black-box models. They can be categorized into gradient-based and perturbation-based approaches. Gradient-based explainers~\cite{selvaraju2017grad,shrikumar2016not,simonyan2013deep, sundararajan2017axiomatic, belharbi2022f} compute attribution scores using the gradients with respect to the inputs or intermediate feature maps. Perturbation-based approaches~\cite{fong2019understanding,fong2017interpretable,ribeiro2016should,zeiler2014visualizing} compute attribution scores via the neural network's response to perturbations. For instance, the Occlusion method~\cite{zeiler2014visualizing} measures feature importance by analyzing changes in the model predictions when applying binary masks to image regions. Some gradient-based and perturbation-based explainability methods use attention weights to generate explanations~\cite{abnar2020quantifying,chefer2021transformer,chefer2021generic}. These methods work only for Transformer-based black-box models and cannot explain the decisions of the state-of-the-art ROP classifier since it is CNN-based.

Recent work on the explainability of ROP classification includes the use of gradient-based explainers like DeepSHAP \cite{wang2021automated} and GradCAM \cite{wu2022development}. However, these generic explainers do not integrate domain-specific image characteristics, such as sparsity and smoothness, which are important for ROP images. 

Several explainability methods have attempted to incorporate sparsity or smoothness or both. The Input$\times$Gradient ~\cite{shrikumar2016not} generates sparse explanations by multiplying the input with the model's output gradient with respect to the input. Nevertheless, it exhibits a lack of smoothness across neighboring vessels, as seen in Figure~\ref{fig:sample figure}(c). Fong \etal ~\cite{fong2019understanding,fong2017interpretable} define their objective function to generate smooth and sparse explanation masks by learning these masks through applying perturbations to images, such as blurring or replacing pixel values with a constant. In Fong \etal ~\cite{fong2017interpretable} the authors incorporate the $\ell_{1}$ norm for sparsity, total variation norm, and upsampling from a low-resolution mask with the application of a 
Gaussian kernel for smoothness. 
While the $\ell_1$ norm indeed leads to sparse solutions, it serves as an approximation of the $\ell_{0}$ norm; therefore, it results in a relaxed sparsity constraint (see also our experiments with $\ell_1$ constraints in Section~\ref{sec:exp}). To leverage a hard sparsity constraint, Fong \etal ~\cite{fong2019understanding} employ a ranking-based area loss and incorporate smoothness using a Gaussian kernel. However, starting from a priori non-smooth masks and smoothing them afterward leads to discontinuous regions and fails to preserve vessel structure, as seen in Figure~\ref{fig:sample figure}(e).

There are multiple adaptations of the method proposed by Fong \etal\cite{fong2019understanding} specifically for the medical domain 
\cite{10.1117/12.2511964, lenis2020domain}. They focus on improving the perturbation method described by Fong \etal \cite{fong2019understanding}, which involves applying blurring or replacing pixel values with a constant value. However, they do not focus on improving the sparsity and smoothness of the explainer.
In contrast, we propose a method that can incorporate sparsity constraints using the $\ell_0$ or $\ell_1$ norms, along with a smoothness constraint during the mask optimization process. Given the optimization problem with multiple constraints, including potentially non-convex constraints, we propose to solve this problem using the Alternating Direction Method of Multipliers (ADMM).

ADMM decomposes problems with combinatorial constraints into subproblems that can be  solved efficiently~\cite{jian2021radio,zhang2018systematic}. Recent studies \cite{takapoui2020simple,leng2018extremely} have demonstrated that ADMM is also effective for solving non-convex problems. This has led to applications in various areas such as weight pruning \cite{zhang2018systematic,jian2021radio,li2019admm}, adversarial attacks \cite{xu2018structured}, and ranking regression \cite{yildiz2020fast}; we expand this repertoire of applications. 


\section{Methodology}
\subsection{Problem formulation}
\label{sec:problem_formulation}
Given a trained neural network model and an ROP image, our goal is to find pixels across the image that affect the model's decision significantly. One way of finding the importance of image pixels is by perturbing the image through masking and observing its effect on the model output.

Formally, a black-box model is defined as a function that maps input image or features to a probability distribution (using a softmax
layer), denoted as $f:\mathbb{R}^{h\times w}\to \mathbb{R}^c,\mathbf{X}\mapsto f(\mathbf{X})$, where $c$ is the number of classes.
We consider an image mask $\mathbf{M} \in \reals_+^{h \times w}$ to be applied to an input image $\mathbf{X} \in \reals^{h \times w}$, where $h$ and $w$ denote the image height and width. We treat this mask as an explanation map: zero values correspond to non-discriminative pixels, while positive values rank remaining pixels with respect to their discriminative value.
To exploit the vessel network structure of segmented fundus images, focusing on vessel features, we only consider non-zero mask elements over image vessels; formally, $\mathbf{M}\in S_0$ where $S_0 = \{\mathbf{M} : \supp(M_{jk}) \subseteq \supp (X_{jk}) \}$, where $\supp(\cdot)$ denotes the support (i.e., the non-zero pixels).
Given such a mask, by taking the Hadamard product of $\mathbf{X}$ and  $\mathbf{M}$, we obtain a \emph{masked image} $\mathbf{X}\odot \mathbf{M} \in \mathbb{R}^{h \times w}$.

Our objective is to create masks $\mathbf{M}$  that highlight important, discriminative pixels across the image and suppress pixels that do not influence model decisions while satisfying two additional desiderata: ensuring mask \emph{sparsity} and \emph{smoothness}. As discussed in Section~\ref{sec:introduction}, sparsity allows us to focus only on important vessel pixels across the image, while smoothness enhances the mask segment contiguity. 
Given $\mathbf{X}$ and a corresponding ground truth label $y$, consider a loss function $l:\mathbb{R}^c \times \mathbb{R} \to \mathbb{R}, f(\mathbf{X}) \times y \mapsto l( f(\mathbf{X}), y)$. We find masks by solving the optimization problem in Eq.~\eqref{eq:masking}:
\begin{equation}
\label{eq:masking}
    \begin{aligned}
    \min_{\mathbf{M}} \quad & l(f(\mathbf{\mathbf{X}\odot \mathbf{M}}),y) + \lambda \nu(\mathbf{M}) \, . \\
     \text{s.t. } \quad & \mathbf{M} \in S_1,\quad \mathbf{M} \in S_2
     \end{aligned}
\end{equation}
Here, the constraint $S_1$ enforces sparsity, $S_2$ enforces that mask values stay within the range $[0,1]$, while the regularization $\nu(\cdot)$ enforces smoothness, and $\lambda\geq 0$ is a regularization parameter. 
In particular, the sparsity constraint is defined as $S_1 = \{\mathbf{M} : \lVert \mathbf{M} \rVert _{p} \leq \alpha_{p}\}\cap S_0$ to limit the number of nonzero pixels, where $\lVert \cdot \rVert_{p}$ denotes the $\ell_p$ norm where $p \in \{0,1\}$, i.e., the number of nonzero elements if $p=0$ and the sum of the absolute values of elements if $p=1$. Here, $\alpha_0 \in \mathbb{N}$ and $\alpha_1 \in \mathbb{R}^{+}$ are constants.
Constraint $S_2$ is defined as $S_2 = \{\mathbf{M} : M_{jk} \in [0,1]\}\cap S_0$. 
Finally, we enforce smoothness using the total variation function~\cite{chambolle2004algorithm,iordache2012total}: $\nu(\mathbf{M}) = \sum_{i} \lVert \mathbf{M}_{i\boldsymbol{\cdot}}-\mathbf{M}_{i-1\boldsymbol{\cdot}} \rVert _{1} + \sum_{j} \lVert \mathbf{M}_{\boldsymbol{\cdot} j}-\mathbf{M}_{\boldsymbol{\cdot} j-1} \rVert _{1} \, ,$
where $\mathbf{M}_{i\boldsymbol{\cdot}}$ and $\mathbf{M}_{\boldsymbol{\cdot} j}$ denote the rows and columns of $\mathbf{M}$.

We follow a \emph{supervised approach} with label $y$ available. However, an unsupervised scenario is possible by using the most likely output class as an objective, or taking the expectation of the loss with respect to the class posterior~\cite{chen2018learning,yoon2018invase}.

\subsection{Computing masks via ADMM}
The optimization problem in Eq.~\eqref{eq:masking} is non-convex with combinatorial constraints. We solve it using the Alternating Direction Method of Multipliers (ADMM), since it allows to decompose the problem into three subproblems that can be solved efficiently and separately \cite{jian2021radio,zhang2018systematic}. For specific non-convex constraints such as the $\ell_{0}$ norm, the ADMM process carries out exactly as the convex case \cite{boyd2011distributed}. Fong et. al.~\cite{fong2017interpretable} have a similar problem definition where they use $\ell_{1}$ norm for sparsity. Nevertheless, $\ell_{1}$ norm serves as an approximation of the $\ell_{0}$ norm. For the weight-pruning problem, Zhang \etal~\cite{zhang2018systematic} demonstrate that employing the $\ell_{0}$ norm with the ADMM process results in sparser solutions compared to $\ell_{1}$ norm-based methods. Using ADMM allows for the use of either $\ell_0$ or $\ell_1$ norms for sparsity constraints, whereas other methods cannot use the $\ell_0$ norm.

We rewrite Eq.~\eqref{eq:masking} with auxiliary variables $\mathbf{M}_1$ and $\mathbf{M}_2$ in Eq.~\eqref{ADMM}, where $g_{S_i}(\xi)$ represent the indicator functions such that $g_{S_i}(\xi)\!=\! 0$ if $\xi\in S_i$ and $g_{S_i}(\xi)\!=\!+\infty$, otherwise.
\begin{equation}
\label{ADMM}
    \begin{aligned}
    &\min_{\mathbf{M}} \; l(\!f(\mathbf{\mathbf{X}\!\!\odot\!\!\mathbf{M}}),\!y) \!+\!\lambda \nu(\mathbf{M}) \!+\!g_{S_1}(\mathbf{M}_1\!)  \!+\! g_{S_2}(\mathbf{M}_2\!) .\\
    &\text{s.t. }  \;  \mathbf{M} = \mathbf{M}_1, \mathbf{M} = \mathbf{M}_2
     \end{aligned}
\end{equation}
Defining dual variables $\mathbf{U}_1, \mathbf{U}_2\in \mathbb{R}^{h \times w}$ and the penalty value $\rho$, we write the augmented Lagrangian problem as:
\begin{equation}
\begin{aligned}
\label{eq:auglagrangian}
    &\mathcal{L}(\mathbf{M}\!, \mathbf{M}_1\!,\mathbf{M}_2\!,\!\mathbf{U}_1\!,\!\mathbf{U}_2\!)\!=\! l(f(\mathbf{\mathbf{X}\!\odot\! \mathbf{M}}),y)\!+\!\lambda \nu(\mathbf{M})\!+\\
    &\!\sum_{i=1}^{2}\!\Big[\!g_{S_i}\!(\mathbf{M}_i) \!\!+\!\! \rho (\text{Tr}(\!\mathbf{U}_i^{T}\!(\!\mathbf{M}\!-\!\mathbf{M}_i\!)) ) \!\!+\!\!\frac{\rho}{2} (\lVert \!\mathbf{M}\!-\! \mathbf{M}_i \!\rVert ^{2}_{2})\!\Big]. 
\end{aligned}
\end{equation}
ADMM algorithm iteratively optimizes the augmented Lagrangian via a primal-dual approach until convergence or a maximum number of iterations. At iteration $k$, the steps are:

\begin{subequations}
\label{eq:ADMM}
    \begin{align}    &\mathbf{M}^{(k)}\!:=\!\argmin_{\mathbf{M}} \mathcal{L}(\mathbf{M}\!,\mathbf{M}_1^{(k\!-\!1)}\!, \mathbf{M}_2^{(k\!-\!1)},\mathbf{U}_1^{(k\!-\!1)}\!,\mathbf{U}_2^{(k\!-\!1)}) , \label{eq:updateM}
     \\
     &\mathbf{M}_1^{(k)}\!,\!\mathbf{M}_2^{(k)}\!:=\!\argmin_{\mathbf{M}_1,\mathbf{M}_2}\mathcal{L}(\mathbf{M}^{(k)}, \mathbf{M}_1, \mathbf{M}_2,\mathbf{U}_1^{(k\!-\!1)},\mathbf{U}_2^{(k\!-\!1)})  ,\label{eq:updateM'}    
    \\
    &\mathbf{U}_i^{(k)}\!:=\! \mathbf{U}_i^{(k\!-\!1)}\!+\! \rho(\mathbf{M}^{(k)}\!-\! \mathbf{M}_i^{(k)}) \text{ for } i=1,2 . \label{eq:updateU}
    \end{align}
\end{subequations}
The problem in Eq.~\eqref{eq:updateM} is equivalent to Eq.~\eqref{eq:updateM_equivalent} and can be solved using standard gradient descent. 
\begin{equation}
    \begin{split}    \label{eq:updateM_equivalent}
    \mathbf{M}^{(k)}\!:=\!& \argmin_{\mathbf{M}} l(f(\mathbf{\mathbf{X}\odot \mathbf{M}}),y) \!+\!\lambda \nu(\mathbf{M}) \\&+ \sum_{i=1}^{2}\frac{\rho}{2} \lVert \mathbf{M} \!-\!\mathbf{M}_i^{(k\!-\!1)}\!+\!\mathbf{U}_i^{(k\!-\!1)}\rVert ^{2}_{2}  \, .
    \end{split}
\end{equation}
Eq.~\eqref{eq:updateM'} for $\mathbf{M}_1$ and $\mathbf{M}_2$ is equivalent to:
\begin{equation}
\label{eq:updateM'_equivalent}
    \mathbf{M}_i^{(k)}\!:=\!\argmin_{\mathbf{M}_i} g_{S_i}(\mathbf{M}_i)\!+\!\frac{\rho}{2} \lVert {\mathbf{M}}^{(k)}\!-\!\mathbf{M}_i\!+\!{\mathbf{U}_i}^{(k\!-\!1)}\rVert ^{2}_{2}\, .
\end{equation}
Crucially, the problems in Eqs.~\eqref{eq:updateM'_equivalent} correspond to Euclidean projection operators that can be solved efficiently in polynomial time~\cite{boyd2011distributed,zhang2018systematic}. Therefore, we define projections for three cases: $S_1$ with the $\ell_0$ norm, $S_1$ with the $\ell_1$ norm, and $S_2$. In particular, we define the Euclidean projection operator as: $\Pi_{S_i}=\argmin_{\mathbf{M}_i \in S_i}\lVert{({\mathbf{M}}^{(k)} + {\mathbf{U}_i}^{(k\!-\!1)}) - \mathbf{M}_i}\rVert ^{2}_{2},$ which maps $({\mathbf{M}}^{(k)} + {\mathbf{U}_i}^{(k\!-\!1)})$ onto the set $S_i$.
When $S_1 = \{\mathbf{M} : \lVert \mathbf{M} \rVert _{0} \leq \alpha_0\}\cap S_0$, where $\alpha_0 \in \mathbb{N}$ is a constant, we update ${\mathbf{M}_1}^{(k)}$ by keeping the top $\alpha_0$ values of $(\mathbf{M}^{(k)} + \mathbf{U}_1^{(k\!-\!1)})$ and setting the rest to zero. We can perform this despite the non-convexity of the set $S_1$ with the $\ell_0$ norm. For $S_1 = \{\mathbf{M} : \lVert \mathbf{M} \rVert _{1} \leq \alpha_1\}\cap S_0$, where $\alpha_1 \in \mathbb{R}^{+}$ is a constant, we follow the projection operation described by Duchi \etal~\cite{duchi2008efficient}. Lastly, for $S_2$, we update ${\mathbf{M}_2}^{(k)}$ by setting the values of $(\mathbf{M}^{(k)} + \mathbf{U}_2^{(k\!-\!1)})$ that are smaller than zero to zero and larger than one to one. 
Algorithm~\ref{alg:algorithm} outlines the ADMM process for computing masks.



\begin{algorithm}[tb]
\caption{SSplain Algorithm}
\label{alg:algorithm}
\textbf{Input}: Black-box model $f(\cdot)$, Input image $\mathbf{X}$, Target label $y$, Loss function $l(\cdot)$, Sparsity parameter $\alpha$, Regularization parameter $\lambda$, Penalty value $\rho$, Maximum iteration $K$\\
\textbf{Output}: Attribution Mask $\mathbf{M}$
\begin{algorithmic}[1] 
\STATE Initialize $\mathbf{M}^{0}$ 
\STATE $\mathbf{U}_i^0$ \!\Assign\! $\mathbf{0}$ \text{ for } $i=1,2$
\STATE $\mathbf{M}_{i}^{0}=\Pi_{S_i} ({\mathbf{M}}^{0} + {\mathbf{U}_i}^{0})$ \text{for} $i=1,2$ 
\FOR{k \Assign 1, \dots, K}
    \STATE $\mathbf{M}^{(k)}$ \!\Assign \!$\argmin_{\mathbf{M}} \big[l(f(\mathbf{\mathbf{X}\odot \mathbf{M}}),y) \!+\!\lambda \nu(\mathbf{M}) \!+\! \frac{\rho}{2}\sum_{i=1}^{2} \lVert \mathbf{M} - \mathbf{M}_i^{(k\!-\!1)}\!+\! \mathbf{U}_i^{(k\!-\!1)}\rVert ^{2}_{2}$\big]
    \STATE $\mathbf{M}_{i}^{(k)}=\Pi_{S_i} ({\mathbf{M}}^{(k)} \!+\! {\mathbf{U}_i}^{(k\!-\!1)} )$ \text{for} $i=1,2$ 
    \STATE $\!\mathbf{U}_i^{(k)}\!$ \!\Assign\! $\!\mathbf{U}_i^{(k\!-\!1)} \!+\! \rho(\!\mathbf{M}^{(k)} \!-\! \mathbf{M}_i^{(k)})$ \text{for} $i=1,2$ 
\ENDFOR
\STATE \textbf{return} $\mathbf{M}^{(K)}$
\end{algorithmic}
\end{algorithm}

\section{Experiments}\label{sec:exp}
\subsection{Dataset and Model Details}
\label{sec:datasetandmodel}
\paragraph{ROP Dataset}
The ROP dataset~\cite{brown2018automated} contains $224\times224$ segmented fundus images. A committee of three experts assigned Reference Standard Diagnostic (RSD) labels~\cite{RSD} as ``Plus'', ``Pre-plus'', or ``Normal'', indicating severe, intermediate, or no ROP, respectively.
For training a classifier, we followed the setup by Brown \etal~\cite{brown2018automated}. We split the data into training and testing sets with 5561 and 100 images, respectively. We use this dataset since a state-of-the-art classification method exists for it, the ROP classification setup proposed by Brown \etal~\cite{brown2018automated}, which achieves a balanced accuracy of 91\%. Following Brown \etal~\cite{brown2018automated}, we train a classifier leveraging the Inception architecture~\cite{szegedy2015going}, while reserving the test set for evaluating different explainers. Note that ROP images in this work are segmented fundus images, which is important since disease severity is determined by the appearance of retinal vessels~\cite{chiang2021international}. 


The focus of our work is not model training. To evaluate SSplain on the ROP dataset, we need a trained model, which we get from the ROP classifier by Brown \etal~\cite{brown2018automated}.

We initialize masks with values proportional to the corresponding pixel intensity between 0 and 1, as follows: $\mathbf{M} = \frac{\mathbf{X}}{\|\mathbf{X}\|_{\infty}}$. The mask initialization depends on the use case. Alternatively, one can initialize masks as all ones assigned to the support of the image using $S_0$, where $M^0_{j,k} = 1$ for $M_{j,k} \in S_0$ with $S_0 = \{\mathbf{M} : \supp(M_{jk}) \subseteq \supp (X_{jk}) \}$. $\supp(\cdot)$ denotes the support (i.e., the non-zero pixels). See Appendix~\ref{sec:mask initialization} for mask initialization comparison. Intensity-based initialization is appropriate for our use case, as the optimization problem is constrained by sparsity and smoothness, and vessel intensity is an appropriate starting point.

We use cross-entropy for $l(\cdot,\cdot)$ and run $50$ iterations of ADMM for each image. We utilize Adam optimizer~\cite{kingma2014adam} to solve problem~\eqref{eq:updateM} with $0.01$ learning rate, $\rho=0.01$ and $\lambda=10^{-5}$. We set $\alpha_0$ in $S_1$ to $50\%$ of $\lVert \mathbf{X} \rVert _{0}$ for the $\ell_0$ norm. To maintain the same sparsity constraints in both the $\ell_0$ and $\ell_1$ norm cases, we use $\alpha_0$ to determine $\alpha_1$. Specifically, when using the $\ell_1$ norm in $S_1$, where $S_1 = \{\mathbf{M} : \lVert \mathbf{M} \rVert _{1} \leq \alpha_1\}\cap S_0$, we set $\alpha_1$ based on $\alpha_0$. At each ADMM iteration, $\alpha_1$ is set to the sum of the top $\alpha_0$ values of $(\mathbf{M}^{(k)} + \mathbf{U}_1^{(k-1)})$. See Appendix~\ref{sec:sparsity alpha} for analyses with different $\alpha$ levels.

\paragraph{Additional Datasets}
We apply SSplain to two other datasets, MNIST~\cite{lecun1998mnist} and FMNIST~\cite{xiao2017fashion}, where images exhibit sparse and smooth structures similar to ROP images. This allows us to evaluate the generality of SSplain.

We train the LeNet-5~\cite{lecun2015lenet} separately on the MNIST and FMNIST datasets using the Adam optimizer~\cite{kingma2014adam} with learning rate 0.001, weight decay 0.0001, for 50 epochs, batch size 32, and early stopping. This results in two separate black-box models, achieving $98.53\%$ test accuracy on MNIST and $88.67\%$ on FMNIST. We then perform our post-hoc explainability analyses on 500 images from the test sets. Refer to Appendix~\ref{sec:additional_experiments} for additional details on the MNIST~\cite{lecun1998mnist} and FMNIST~\cite{xiao2017fashion} datasets and model details.

\subsection{Competing Methods}
\label{sec:competing_methods}
We compare SSplain with nine state-of-the-art explainers available within TorchRay~\cite{fong2019understanding} and Captum~\cite{kokhlikyan2020captum} libraries. Here, we provide the application details for the ROP dataset.
\sloppy
\textit{Gradient-based:}  Saliency~\cite{simonyan2013deep}, Guided Grad-CAM~\cite{selvaraju2017grad}, Integrated Gradients~\cite{sundararajan2017axiomatic} and DeepSHAP~\cite{lundberg2017unified}. We apply Guided Grad-CAM method to the output of the last Inception layer (``inception 5b'').
\textit{Perturbation-based:} KernelSHAP~\cite{lundberg2017unified}, LIME~\cite{ribeiro2016should} and Occlusion ~\cite{zeiler2014visualizing}. LIME and KernelSHAP employ $200$ input samples. Occlusion uses $16\times16$ windows with $4\times4$ strides.
\textit{Incorporating sparsity or smoothness or both:} Extremal Perturbation ~\cite{fong2019understanding} and Input$\times$
Gradient ~\cite{shrikumar2016not}. Extremal Perturbation uses a Gaussian blur, a ``hybrid'' loss with $400$ iterations, $0.05$ smoothing factor, and a target area to $50\%$ of $\lVert \mathbf{X} \rVert _{0}$. It initializes mask values to all ones. See Appendix~\ref{sec:competing method details} for method details, and Appendix~\ref{sec:additional_experiments} for the application to MNIST and FMNIST.

\begin{figure*}[t!]
    \centering    
     \begin{subfigure}[b]{1.0\textwidth}
         \centering
        \includegraphics[width=1.0\linewidth]{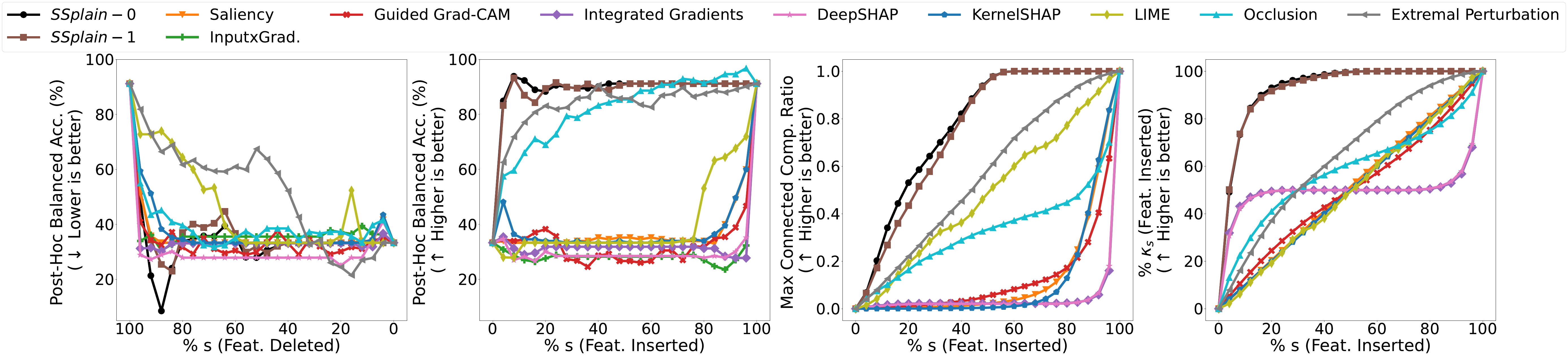}
         \caption{Accuracy, connected components, normalized sparsity $\kappa_s$ metrics vs.~sparsity $s$.}
         \label{fig:results_level_s}
     \end{subfigure}
      \begin{subfigure}[b]{1.0\textwidth}
         \centering         \includegraphics[width=1.0\linewidth]{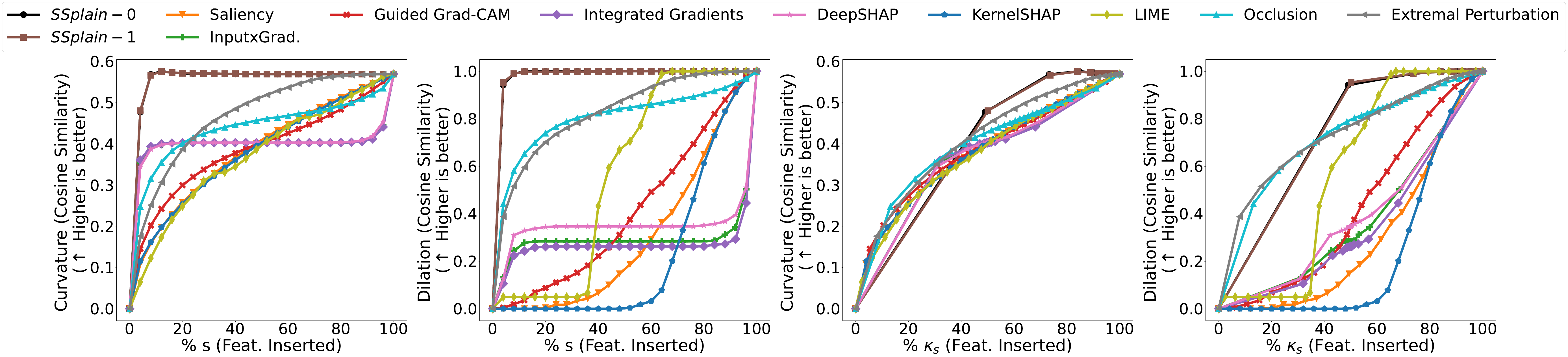}
         \caption{Curvature and dilation analyses vs.~sparsity $s$ and~normalized sparsity $\kappa_s$.}
         \label{fig:results_s_norm}
     \end{subfigure}
    \caption{Comparison of explainers for the ROP dataset: SSplain-0 ($S_1$ with $\ell_0$ constraint), SSplain-1 ($S_1$ with $\ell_1$ constraint), Saliency~\cite{simonyan2013deep}, Input$\times$Gradient~\cite{shrikumar2016not}, Guided Grad-CAM~\cite{selvaraju2017grad},  Integrated Gradients~\cite{sundararajan2017axiomatic}, DeepSHAP~\cite{lundberg2017unified}, KernelSHAP~\cite{lundberg2017unified}, LIME~\cite{ribeiro2016should}, Occlusion~\cite{zeiler2014visualizing} and Extremal Perturbation~\cite{fong2019understanding}. (a) We report the average of: Post-hoc balanced accuracy with deletion (lower is better) and insertion (higher is better) of pixels with the highest attribution scores, connected components ratio during the insertion process, and sparsity $s$ versus normalized sparsity $\kappa_s$ during the insertion process. (b) Curvature and dilation similarity with insertion process with respect to sparsity $s$ and normalized sparsity $\kappa_s$. In (a), both SSplain methods consistently outperform competitors and competitors give importance to the background. In (b), both at the same $s$ and $\kappa_s$ level, SSplain methods exhibit higher curvature and dilation similarity.}
    \label{fig:methods}
\end{figure*}
\begin{figure*}[t]
    \centering    \includegraphics[width=1.0\linewidth]{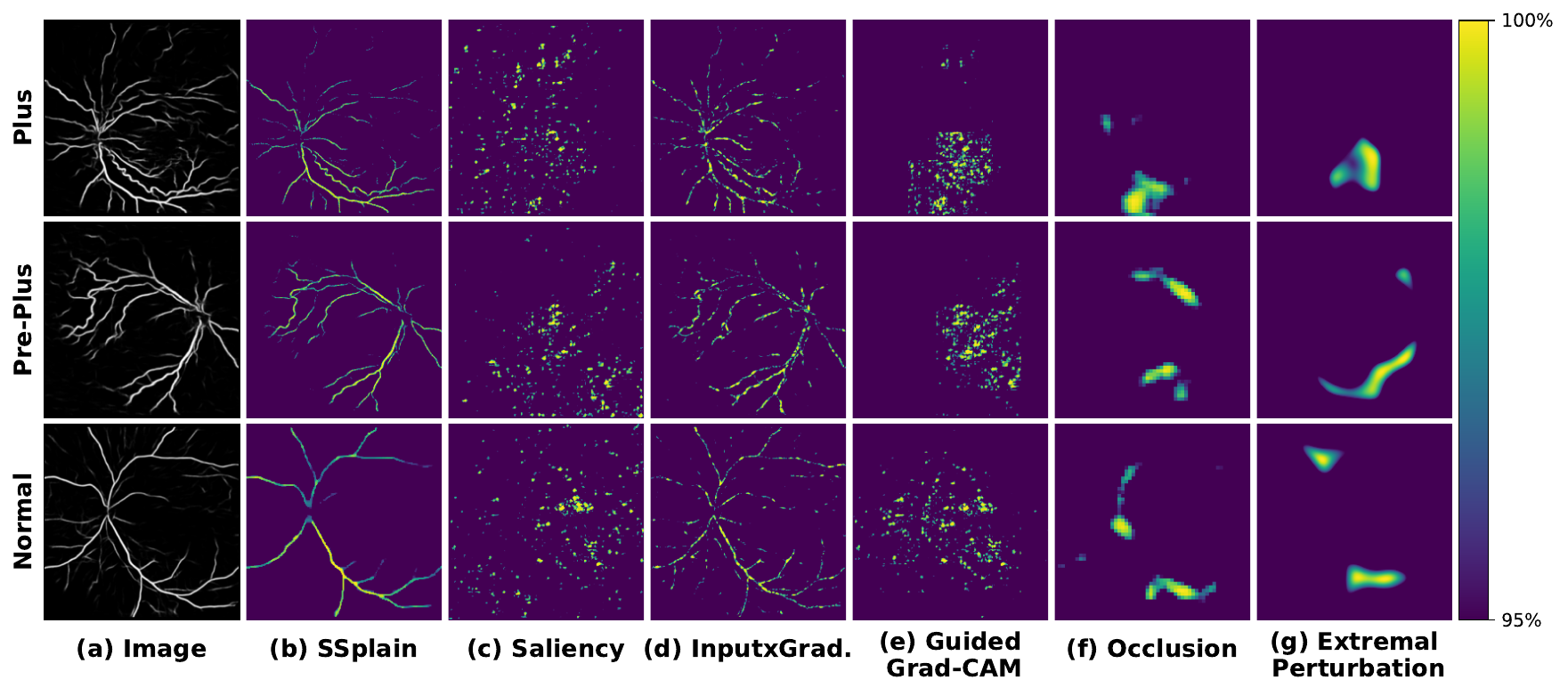}
    \caption{Comparison with other explainers for the ROP dataset. From left to right: (a) images, explanation maps generated using (b) SSplain-0, (c) Saliency~\cite{simonyan2013deep}, (d) Input$\times$Gradient~\cite{shrikumar2016not}, (e) Guided Grad-CAM~\cite{selvaraju2017grad}, (f) Occlusion~\cite{zeiler2014visualizing} and (g) Extremal Perturbation~\cite{fong2019understanding}. We visualize only the top 5\% of pixels, as ranked by the corresponding explanation maps, coloring the remaining pixels as purple. SSplain effectively preserves image structures and assigns more importance to tortuous and dilated regions. Except Input$\times$Gradient, competitors treat images as a whole and assign importance to pixels that are not associated with vessels.}
    \label{fig:methods_maps}
\end{figure*}

\subsection{Experimental Setup}  
\paragraph{Insertion and Deletion Tests}
We follow a supervised approach to generate explanations, using true labels as targets. This is because we evaluate explainer performance not only based on accuracy but also on domain-specific feature similarity for ROP, which are disease and severity indicators. For consistency, we use this approach throughout the paper. See Appendix~\ref{sec:unsupervised_generation} for results using an unsupervised setup, using the model predictions as targets. Subsequently, we evaluate several metrics defined in Section~\ref{sec:metrics} using the insertion and deletion processes described by Petsiuk \etal~\cite{petsiuk2018rise}. The insertion/deletion process evaluates performance by adding/ removing image pixels with the highest attribution scores. We evaluate this process with respect to the sparsity level $s$, representing the fraction of pixels selected between $0\%$ and $100\%$. At each $s$, we obtain a corresponding image $\mathbf{X}_s$. Some methods pick black pixels, i.e., the background. Therefore, we also evaluate performance with respect to the normalized sparsity $\kappa_s = \frac{\lVert \mathbf{X}_s \rVert _{1}}{\lVert \mathbf{X} \rVert _{1}}$.
\paragraph{Sanity Checks}
We evaluate the sensitivity of SSplain using 
 ``Model Parameter Randomization Test'' by Adebayo \etal ~\cite{adebayo2018sanity}. We recursively randomize the model weights, layer by layer from top to bottom, while the explainer generates attribution scores for each case. This test assesses explainer sensitivity to model changes, and explainers should produce different attribution scores with randomized weights.
 
\paragraph{Computing Infrastructure} We ran experiments using PyTorch~\cite{paszke2019pytorch} on an Intel Core i7-9700K CPU with 64GB RAM.

\paragraph{Computational Costs}
SSplain’s execution time depends on the number of ADMM iterations, image size, and size of the black-box model. At each iteration, we perform a standard gradient descent (line 5 of Algorithm~\ref{alg:algorithm}) and projection operations (line 6 of Algorithm~\ref{alg:algorithm}). These projection operations are polynomial in time with respect to the image size, where we select the top $\alpha$ values for the $\ell_0$ norm projection ($S_1$ constraint) and apply clipping for the $S_2$ constraint.

SSplain’s runtime is comparable with the perturbation methods, the explainability family to which it belongs. Extremal Perturbation's runtime depends on the iteration number, image size, patch size, and model size. KernelSHAP and LIME depend on the model size and the perturbation samples. Occlusion depends on the model size, image size, patch size, and stride. Gradient-based methods depend on model size and are typically faster than input-perturbation methods. Note that explanations and runtimes are sensitive to method configurations. Please see Appendix~\ref{sec:execution time} for the execution times.

\label{sec:experiments metrics analyses}
\subsection{Evaluation Metrics}
\label{sec:metrics}
We evaluate explainers' performance on balanced accuracy. We use ``Connected Components'' to evaluate smoothness on ROP data. We also define two domain-related metrics to measure the explainer's ability to capture semantically meaningful ROP features: \emph{tortuosity} and \emph{dilation}.

\paragraph{Connected components}
We measure the smoothness of explanation maps, focusing particularly on the smoothness of vessel structures at each insertion step $s$. To achieve this, we extract graphs from $\mathbf{X}$ and $\mathbf{X}_s$, and identify the length of the largest connected component in each graph. This length is the number of nodes in the largest connected component. To both extract graphs and find the connected components in graphs, we use the NetworkX library~\cite{hagberg2008exploring}; we describe processes in Appendix~\ref{sec:supp_connected_components}. We compute the ratio of the length of the largest connected component in $\mathbf{X}_s$ to the length of the largest connected component in $\mathbf{X}$. A higher largest connected component value at $s$ indicates that the inserted pixels create smoother vessel structures.
\paragraph{Curvature} 
Tortuosity is an ROP indicator. Therefore, we also evaluate the explainers' ability to capture tortuous regions at each insertion step $s$. For this, we measure the similarity of images at each insertion step $s$ with pixel-wise curvature values $\mathbf{C}\in \reals_+^{h \times w}$ for $\mathbf{X}$. These curvature values are calculated using the Tubular Curvature Filter Method~\cite{sunger2023tubular}. Higher pixel-wise curvature values indicate greater tortuosity. Refer to the Appendix~\ref{sec:supp_curvature} for details on this metric.

We use $\text{cos}(\mathbf{X}_s,\mathbf{C}) = \frac{\mathbf{X}_s \cdot \mathbf{C}}{\rVert\mathbf{X}_s\lVert \rVert\mathbf{C}\lVert}$ to compute the cosine similarity between the curvature values $\mathbf{C}$ and $\mathbf{X}_s$. A higher value at $s$ shows that the inserted pixels capture tortuous regions, and doing this at an earlier $s$ indicates that the explainer assigns greater importance to these regions.

\paragraph{Dilation}
Another disease indicator is dilation, i.e., dilated vessels on the retina. To assess if explainers identify dilated vessels, we measure the similarity between dilated vessels in $\mathbf{X}$ and $\mathbf{X}_s$ during the insertion process. We extract dilation using the Average Segment Diameter (ASD)~\cite{ataer2015computer}. ASD measures the total number of pixels in a vessel branch divided by the curve length, allowing us to evaluate how well the explainer captures dilation at each insertion step $s$.

We evaluate $\text{cos}(\mathbf{D_s},\mathbf{D})$, where $\mathbf{D}\in \mathbb{R}^{b\times 1}$ contains the ASD features from $\mathbf{X}$ and $\mathbf{D}_s$ from  $\mathbf{X}_s$. $b$ is the number of vessel branches. Similar to the curvature metric, a higher value at an earlier $s$ indicates that the explainer assigns greater importance to dilated vessel regions. We extract ASD features following Ataer \etal~\cite{ataer2015computer} and Tian \etal~\cite{tian2019severity}. See Appendix~\ref{sec:supp_dilation} for details.

\subsection{Post-hoc and Similarity Analyses}
\label{sec:posthoc}
We report averages of the metrics defined above over the test set for ROP dataset. 
SSplain-0 refers to when $S_1$ includes the $\ell_0$ norm as the sparsity constraint, while SSplain-1 refers to when it includes the $\ell_1$ norm.
Figure~\ref{fig:methods} shows the evaluation of explainability methods. Across all analyses from Figure~\ref{fig:results_level_s}, SSplain with both $\ell_0$ and $\ell_1$ consistently outperforms comparative methods, demonstrating lower deletion and higher insertion post-hoc accuracy, as well as higher values for max connected components ratio. 
Across all analyses, SSplain-0 performs as well as or better than SSplain-1. Thus, we include only SSplain-0 in the following analyses.

Except for the deletion analysis, the third-best is the Extremal Perturbation. The last column of Figure~\ref{fig:results_level_s} shows competitors assign importance to the background, requiring more steps to reach the $\kappa_s$ level that SSplain achieves faster. Figure~\ref{fig:results_s_norm} shows that at the same $s$ and $\kappa_s$ level, \emph{SSplain exhibits higher curvature and dilation similarity}, indicating better capture of domain-related features. See Appendix~\ref{sec:ablation study} for an ablation study on the ROP dataset. 

\begin{figure}[h!]
    \centering
    \includegraphics[width=1.0\linewidth]{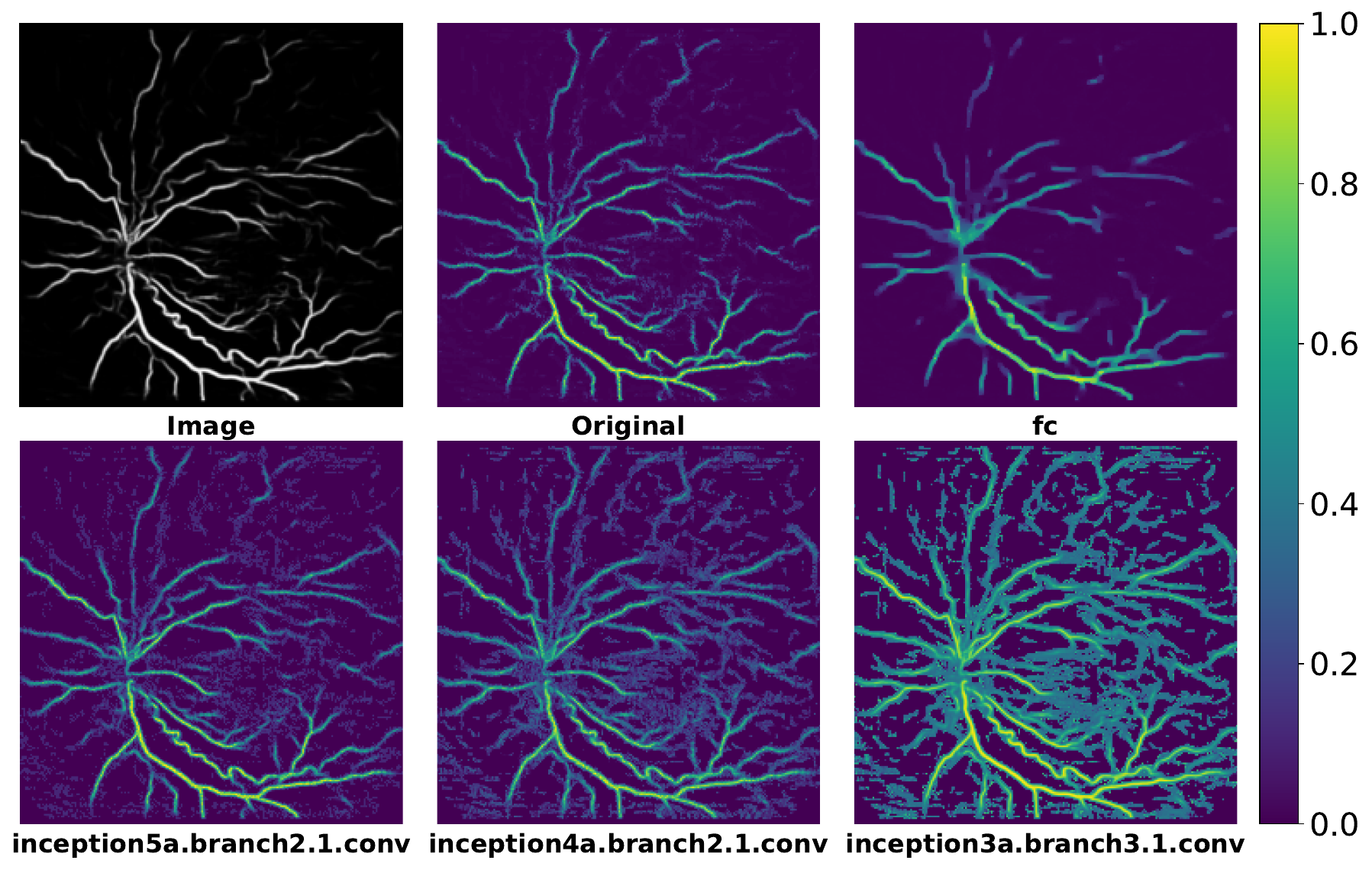}
    \caption{Sanity check~\cite{adebayo2018sanity} for an ROP image. Model weights are progressively randomized from the ``fc'' layer to the ``inception3a.branch3.1.conv'' layer in the Inception model~\cite{szegedy2015going}. The first row shows a sample image, original attribution scores, and attribution scores with only the ``fc'' layer randomized. The second row provides examples from the other layers during progressive randomization. SSplain is sensitive to model weights, i.e., attribution scores change when we randomize the weights as desired. $S_0$ ensures the mask applies only to the vessels, so the mask values for the background do not change.
    }
    \label{fig:sanity_check_with_S0}
\end{figure}

\begin{figure}[t]
    \centering    
    \begin{subfigure}[b]{0.47\textwidth}
    \centering    
    \includegraphics[width=1.0\linewidth]{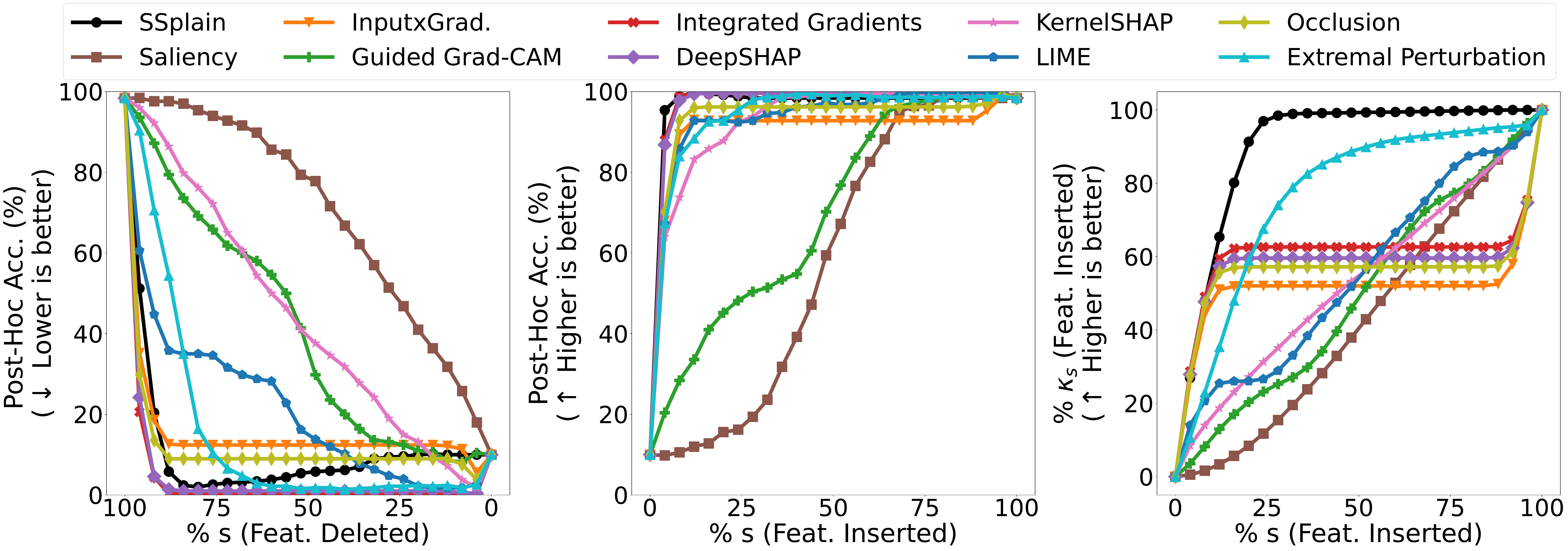}
     \label{fig:imagenet_main}
    \caption{MNIST results.}
    \label{fig:mnist study}
    \end{subfigure}
    \begin{subfigure}[b]{0.47\textwidth}
    \centering    
    \includegraphics[width=1.0\linewidth]{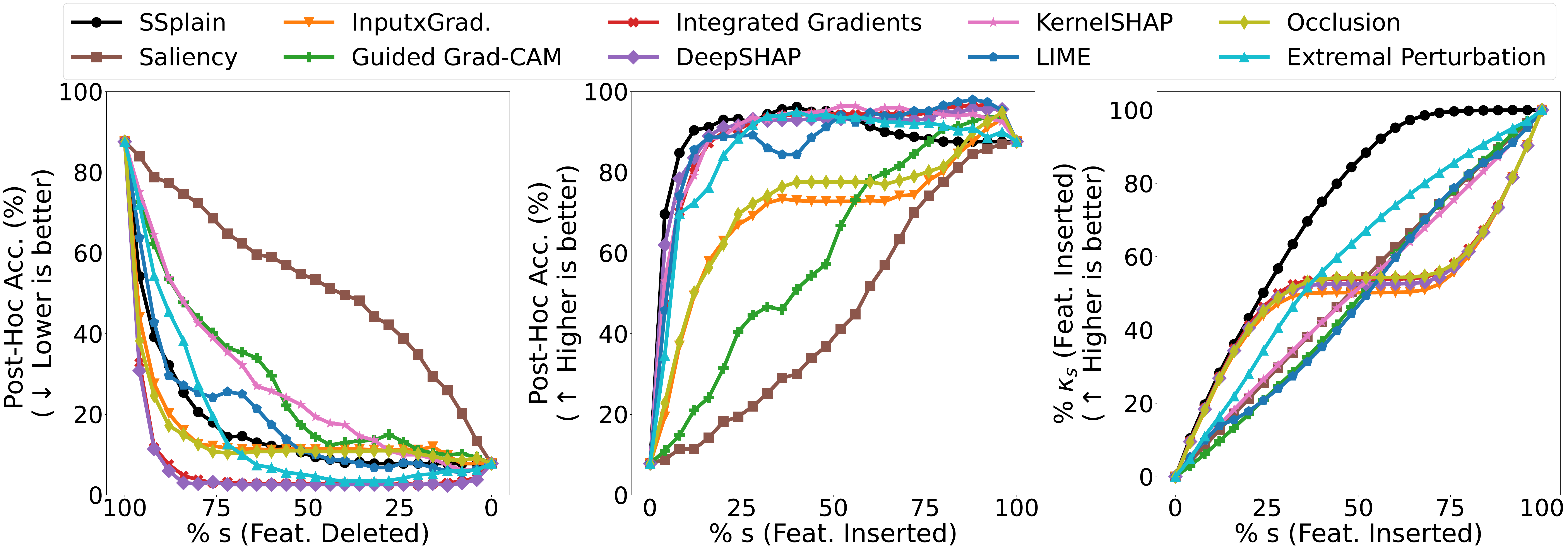}
    \caption{FMNIST results.}
    \label{fig:fmnist study}
    \end{subfigure}
    \caption{Comparison of explainers: SSplain-0, Saliency~\cite{simonyan2013deep}, Input$\times$Gradient~\cite{shrikumar2016not}, Guided Grad-CAM~\cite{selvaraju2017grad}, Integrated Gradients~\cite{sundararajan2017axiomatic}, DeepSHAP~\cite{lundberg2017unified}, KernelSHAP~\cite{lundberg2017unified}, LIME~\cite{ribeiro2016should}, Occlusion~\cite{zeiler2014visualizing} and Extremal Perturbation~\cite{fong2019understanding} for (a) MNIST and (b) FMNIST datasets. From left to right, we report the average of: Post-hoc accuracy with deletion (lower is better) and insertion (higher is better) of pixels with the highest attribution scores, and sparsity $s$ vs normalized sparsity $\kappa_s$ during the insertion process. For MNIST, SSplain outperforms the competing methods in all analyses. It also stabilizes earlier than other methods, meaning that performance changes become minimal when inserting or deleting pixels in later attempts. For FMNIST, we observe that SSplain performs best in accuracy analysis with the insertion process and in normalized sparsity $\kappa_s$ analysis.  Input$\times$Gradient and Occlusion outperform SSplain in deletion analyses. Despite their superior performance in deletion analyses, these methods significantly underperform compared to SSplain in insertion analysis, having lower insertion accuracy.}
\end{figure}

\subsection{Visualization}

Visualizing the most important features can be challenging when comparing the outputs of the different methodologies and objectives (see Appendix~\ref{sec:supplemet visualization rop}). To circumvent this issue, we visualize only the top 5\% of pixels, as ranked by the corresponding explanation maps, coloring the remaining pixels purple. 
We illustrate this visualization procedure for the examples from the ROP dataset in Figure~\ref{fig:methods_maps}. 
Here, we only include SSplain with the $\ell_0$ norm used in $S_1$.
SSplain preserves image structures and assigns more importance to pixels with high tortuosity and dilation while maintaining smoothness. The $S_0$ and $S_1$ constraints in SSplain’s optimization provide this sparsity advantage. Among competitors, only Input$\times$Gradient has an $S_0$-like component, restricting explanations by multiplying the input with the output gradient. While Input$\times$Gradient also preserves vessel shapes, it fails to generate smooth attribution scores and capture tortuous and dilated areas. SSplain generates attribution scores specifically for pixels associated with vessels by preserving their structure and enforcing sparsity. In contrast, competitors treat images as a whole and thus do not distinctly focus on the pixels associated with vessels. 
See Appendix~\ref{sec:S_1 with l0 and l1} for visualizations of SSplain-0 and SSplain-1. Refer to Appendix~\ref{sec:ssplain vs extremal perturbations} for SSplain-0 and Extremal Perturbations (pixel-wise), which uses the $\ell_1$ norm to illustrate an example where both methods share an all-ones mask initialization and SSplain is without $S_0$.

\paragraph{Sanity Check}
In Figure~\ref{fig:sanity_check_with_S0}, we visualize the sensitivity of the attribution scores by progressively randomizing the model weights from top to bottom, as proposed by Adebayo \etal ~\cite{adebayo2018sanity}, and apply SSplain-0 to the ROP dataset. This visualization shows how sensitive attribution scores are to changes in model weights. It is desired that the scores vary with randomized weights. We observe these changes in Figure~\ref{fig:sanity_check_with_S0} that the attribution scores change with the randomized weights. Note that $S_0$ ensures the mask applies only to the vessels, so background mask values do not change.

\subsection{Analyses on the Additional Datasets}
\label{sec:add_posthoc}
\paragraph{MNIST}
Figure~\ref{fig:mnist study} demonstrates the performance of explainability methods across different analyses and metrics for the MNIST dataset. SSplain-0 consistently outperforms the competing methods, demonstrating lower deletion and higher insertion post-hoc accuracy, as well as higher normalized sparsity $\kappa_s$. Additionally, SSplain stabilizes early, reaching its maximum sooner and maintaining stability until the end of the insertion process. Similar to ROP results, competitors prioritize both background and non-zero pixels, requiring more steps to achieve the $\kappa_s$ level that SSplain reaches in just a few steps. Refer to Appendix~\ref{sec:vizualizationsforMNISTandFMNIST} for attribution score visualizations of the MNIST dataset.

\paragraph{FMNIST}
Figure~\ref{fig:fmnist study} shows the performance of explainers across different analyses and metrics for the FMNIST dataset. Although SSplain-0 excels in insertion-based accuracy analysis and in normalized sparsity $\kappa_s$ analysis, other methods such as Occlusion and Input$\times$Gradient perform better than SSplain in deletion analysis. These methods, however, significantly underperform compared to SSplain in insertion analyses, exhibiting lower insertion accuracy. This means that the pixels with the highest attribution scores in the Occlusion and Input$\times$Gradient methods significantly affect the classification when removed from the image but do not impact it when added. Similar to the MNIST and ROP results, SSplain stabilizes earlier. See Appendix~\ref{sec:vizualizationsforMNISTandFMNIST} for attribution score visualizations of the FMNIST dataset.

\section{Conclusion}
\label{sec:conclusion}
This paper proposes a black-box explainer designed to preserve input image structures by enhancing smoothness and sparsity for ROP classification. Unlike previous methods, SSplain effectively maintains input image structures, such as smoothness and sparsity, thus it generates realistic explanations to enhance clinicians' trust in model outputs. To generate attribution scores while ensuring both sparsity and smoothness, we define and solve a non-convex optimization problem with combinatorial constraints using ADMM. Experiments demonstrate that SSplain outperforms nine commonly used explainers not only in post-hoc performance but also in domain-based feature similarity analyses of tortuosity and dilation. Additionally, the sanity check visualizations show that SSplain is sensitive to changes in the model, which is desired for explainers.

In addition to showing SSplain's performance on the ROP dataset, our experiments include additional results on the MNIST and FMNIST datasets to demonstrate SSplain’s applicability to other domains.  SSplain is specifically defined for segmented ROP images, which have sparse and smooth characteristics. Therefore, it is also dependent on the segmentation quality in terms of smoothness and sparsity. While we define the optimization problem specifically for this, for other possible problem formulations requiring different optimization constraints, one can add or substitute constraints based on domain knowledge and solve them using ADMM.

\section*{Acknowledgments}
This work was supported by grants R01 EY019474, and P30 EY10572 from the National Institutes of Health, by unrestricted departmental funding from Research to Prevent Blindness, and by the Malcolm Marquis Innovation Fund.

{
    \small
    \bibliographystyle{ieeenat_fullname}
    \bibliography{wacv}

@String(ECCV= {Eur. Conf. Comput. Vis.})

@String(AAAI = {AAAI})

@String(ECCV  = {ECCV})

@article{iordache2012total,
  title={{Total variation spatial regularization for sparse hyperspectral unmixing}},
  author={Iordache, Marian-Daniel and Bioucas-Dias, Jos{\'e} M and Plaza, Antonio},
  journal={IEEE Transactions on Geoscience and Remote Sensing},
  volume={50},
  number={11},
  pages={4484--4502},
  year={2012},
  publisher={IEEE}
}

@article{chambolle2004algorithm,
  title={{An algorithm for total variation minimization and applications}},
  author={Chambolle, Antonin},
  journal={Journal of Mathematical imaging and vision},
  volume={20},
  pages={89--97},
  year={2004},
  publisher={Springer}
}

@article{jian2021radio,
  title={{Radio frequency fingerprinting on the edge}},
  author={Jian, Tong and Gong, Yifan and Zhan, Zheng and Shi, Runbin and Soltani, Nasim and Wang, Zifeng and Dy, Jennifer and Chowdhury, Kaushik and Wang, Yanzhi and Ioannidis, Stratis},
  journal={IEEE Transactions on Mobile Computing},
  volume={21},
  number={11},
  pages={4078--4093},
  year={2021},
  publisher={IEEE}
}

@inproceedings{zhang2018systematic,
  title={{A systematic dnn weight pruning framework using alternating direction method of multipliers}},
  author={Zhang, Tianyun and Ye, Shaokai and Zhang, Kaiqi and Tang, Jian and Wen, Wujie and Fardad, Makan and Wang, Yanzhi},
  booktitle={Proceedings of the European conference on computer vision (ECCV)},
  pages={184--199},
  year={2018}
}

@inproceedings{fong2019understanding,
  title={Understanding deep networks via extremal perturbations and smooth masks},
  author={Fong, Ruth and Patrick, Mandela and Vedaldi, Andrea},
  booktitle={Proceedings of the IEEE/CVF international conference on computer vision},
  pages={2950--2958},
  year={2019}
}

@inproceedings{fong2017interpretable,
  title={Interpretable explanations of black boxes by meaningful perturbation},
  author={Fong, Ruth C and Vedaldi, Andrea},
  booktitle={Proceedings of the IEEE international conference on computer vision},
  pages={3429--3437},
  year={2017}
}

@article{brown2018automated,
  title={Automated diagnosis of plus disease in retinopathy of prematurity using deep convolutional neural networks},
  author={Brown, James M and Campbell, J Peter and Beers, Andrew and Chang, Ken and Ostmo, Susan and Chan, RV Paul and Dy, Jennifer and Erdogmus, Deniz and Ioannidis, Stratis and Kalpathy-Cramer, Jayashree and others},
  journal={JAMA ophthalmology},
  volume={136},
  number={7},
  pages={803--810},
  year={2018},
  publisher={American Medical Association}
}

@article{RSD,
author = {Ryan, Michael and Ostmo, Susan and Jonas, Karyn and Berrocal, Audina and Drenser, Kimberly and Horowitz, Jason and Lee, Thomas and Simmons, Charles and Martínez-Castellanos, María and Chan, Robison and Chiang, Michael},
year = {2014},
month = {11},
pages = {1902-10},
title = {Development and Evaluation of Reference Standards for Image-based Telemedicine Diagnosis and Clinical Research Studies in Ophthalmology},
volume = {2014},
journal = {AMIA ... Annual Symposium proceedings / AMIA Symposium. AMIA Symposium}
}

@inproceedings{szegedy2015going,
  title={Going deeper with convolutions},
  author={Szegedy, Christian and Liu, Wei and Jia, Yangqing and Sermanet, Pierre and Reed, Scott and Anguelov, Dragomir and Erhan, Dumitru and Vanhoucke, Vincent and Rabinovich, Andrew},
  booktitle={Proceedings of the IEEE conference on computer vision and pattern recognition},
  pages={1--9},
  year={2015}
}

@article{petsiuk2018rise,
  title={Rise: Randomized input sampling for explanation of black-box models},
  author={Petsiuk, Vitali and Das, Abir and Saenko, Kate},
  journal={arXiv preprint arXiv:1806.07421},
  year={2018}
}

@article{sunger2023tubular,
  title={Tubular Curvature Filter: Implicit Pointwise Curvature Calculation Method for Tubular Objects},
  author={Sunger, Elifnur and Kalkanli, Beyza and Yildiz, Veysi and Imbiriba, Tales and Campbell, Peter and Erdogmus, Deniz},
  journal={arXiv preprint arXiv:2311.11931},
  year={2023}
}

@article{ataer2015computer,
  title={Computer-based image analysis for plus disease diagnosis in retinopathy of prematurity: performance of the “i-ROP” system and image features associated with expert diagnosis},
  author={Ataer-Cansizoglu, Esra and Bolon-Canedo, Veronica and Campbell, J Peter and Bozkurt, Alican and Erdogmus, Deniz and Kalpathy-Cramer, Jayashree and Patel, Samir and Jonas, Karyn and Chan, RV Paul and Ostmo, Susan and others},
  journal={Translational vision science \& technology},
  volume={4},
  number={6},
  pages={5--5},
  year={2015},
  publisher={The Association for Research in Vision and Ophthalmology}
}

@article{kokhlikyan2020captum,
  title={Captum: A unified and generic model interpretability library for pytorch},
  author={Kokhlikyan, Narine and Miglani, Vivek and Martin, Miguel and Wang, Edward and Alsallakh, Bilal and Reynolds, Jonathan and Melnikov, Alexander and Kliushkina, Natalia and Araya, Carlos and Yan, Siqi and others},
  journal={arXiv preprint arXiv:2009.07896},
  year={2020}
}

@article{kingma2014adam,
  title={Adam: A method for stochastic optimization},
  author={Kingma, Diederik P and Ba, Jimmy},
  journal={arXiv preprint arXiv:1412.6980},
  year={2014}
}

@article{simonyan2013deep,
  title={Deep inside convolutional networks: Visualising image classification models and saliency maps},
  author={Simonyan, Karen and Vedaldi, Andrea and Zisserman, Andrew},
  journal={arXiv preprint arXiv:1312.6034},
  year={2013}
}

@article{shrikumar2016not,
  title={Not just a black box: Learning important features through propagating activation differences},
  author={Shrikumar, Avanti and Greenside, Peyton and Shcherbina, Anna and Kundaje, Anshul},
  journal={arXiv preprint arXiv:1605.01713},
  year={2016}
}

@inproceedings{selvaraju2017grad,
  title={Grad-cam: Visual explanations from deep networks via gradient-based localization},
  author={Selvaraju, Ramprasaath R and Cogswell, Michael and Das, Abhishek and Vedantam, Ramakrishna and Parikh, Devi and Batra, Dhruv},
  booktitle={Proceedings of the IEEE international conference on computer vision},
  pages={618--626},
  year={2017}
}

@article{lundberg2017unified,
  title={A unified approach to interpreting model predictions},
  author={Lundberg, Scott M and Lee, Su-In},
  journal={Advances in neural information processing systems},
  volume={30},
  year={2017}
}

@inproceedings{ribeiro2016should,
  title={" Why should i trust you?" Explaining the predictions of any classifier},
  author={Ribeiro, Marco Tulio and Singh, Sameer and Guestrin, Carlos},
  booktitle={Proceedings of the 22nd ACM SIGKDD international conference on knowledge discovery and data mining},
  pages={1135--1144},
  year={2016}
}

@inproceedings{zeiler2014visualizing,
  title={Visualizing and understanding convolutional networks},
  author={Zeiler, Matthew D and Fergus, Rob},
  booktitle={Computer Vision--ECCV 2014: 13th European Conference, Zurich, Switzerland, September 6-12, 2014, Proceedings, Part I 13},
  pages={818--833},
  year={2014},
  organization={Springer}
}

@article{boyd2011distributed,
  title={Distributed optimization and statistical learning via the alternating direction method of multipliers},
  author={Boyd, Stephen and Parikh, Neal and Chu, Eric and Peleato, Borja and Eckstein, Jonathan and others},
  journal={Foundations and Trends{\textregistered} in Machine learning},
  volume={3},
  number={1},
  pages={1--122},
  year={2011},
  publisher={Now Publishers, Inc.}
}

@techreport{hagberg2008exploring,
  title={Exploring network structure, dynamics, and function using NetworkX},
  author={Hagberg, Aric and Swart, Pieter and S Chult, Daniel},
  year={2008},
  institution={Los Alamos National Lab.(LANL), Los Alamos, NM (United States)}
}

@inproceedings{worrall2016automated,
  title={Automated retinopathy of prematurity case detection with convolutional neural networks},
  author={Worrall, Daniel E and Wilson, Clare M and Brostow, Gabriel J},
  booktitle={International Workshop on Deep Learning in Medical Image Analysis},
  pages={68--76},
  year={2016},
  organization={Springer}
}

@inproceedings{yildiz2021structurally,
  title={Structurally Guided Channel Attention Networks: SGCA-Net},
  author={Yildiz, Veysi and Dy, Jennifer and Campbell, Peter and Ostmo, Susan and Chiang, Michael and Ioannidis, Stratis and Erdogmus, Deniz},
  booktitle={The 14th PErvasive Technologies Related to Assistive Environments Conference},
  pages={93--96},
  year={2021}
}

@inproceedings{yildiz2021structural,
  title={Structural visual guidance attention networks in retinopathy of prematurity},
  author={Yildiz, Veysi and Ioannidis, Stratis and Yildiz, Ilkay and Tian, Peng and Campbell, John Peter and Ostmo, Susan and Kalpathy-Cramer, Jayashree and Chiang, Michael F and Erdo{\u{g}}mu{\c{s}}, D and Dy, J},
  booktitle={2021 IEEE 18th International Symposium on Biomedical Imaging (ISBI)},
  pages={353--357},
  year={2021},
  organization={IEEE}
}

@article{chiang2021international,
  title={International classification of retinopathy of prematurity},
  author={Chiang, Michael F and Quinn, Graham E and Fielder, Alistair R and Ostmo, Susan R and Chan, RV Paul and Berrocal, Audina and Binenbaum, Gil and Blair, Michael and Campbell, J Peter and Capone Jr, Antonio and others},
  journal={Ophthalmology},
  volume={128},
  number={10},
  pages={e51--e68},
  year={2021},
  publisher={Elsevier}
}

@article{wang2021automated,
  title={Automated explainable multidimensional deep learning platform of retinal images for retinopathy of prematurity screening},
  author={Wang, Ji and Ji, Jie and Zhang, Mingzhi and Lin, Jian-Wei and Zhang, Guihua and Gong, Weifen and Cen, Ling-Ping and Lu, Yamei and Huang, Xuelin and Huang, Dingguo and others},
  journal={JAMA network open},
  volume={4},
  number={5},
  pages={e218758--e218758},
  year={2021},
  publisher={American Medical Association}
}

@article{wu2022development,
  title={Development and validation of a deep learning model to predict the occurrence and severity of retinopathy of prematurity},
  author={Wu, Qiaowei and Hu, Yijun and Mo, Zhenyao and Wu, Rong and Zhang, Xiayin and Yang, Yahan and Liu, Baoyi and Xiao, Yu and Zeng, Xiaomin and Lin, Zhanjie and others},
  journal={JAMA Network Open},
  volume={5},
  number={6},
  pages={e2217447--e2217447},
  year={2022},
  publisher={American Medical Association}
}

@article{gilbert2001childhood,
  title={Childhood blindness in the context of VISION 2020: the right to sight},
  author={Gilbert, Clare and Foster, Allen},
  journal={Bulletin of the World Health Organization},
  volume={79},
  number={3},
  pages={227--232},
  year={2001},
  publisher={SciELO Public Health}
}

@article{hanif2021applications,
  title={Applications of interpretability in deep learning models for ophthalmology},
  author={Hanif, Adam M and Beqiri, Sara and Keane, Pearse A and Campbell, J Peter},
  journal={Current opinion in ophthalmology},
  volume={32},
  number={5},
  pages={452},
  year={2021},
  publisher={NIH Public Access}
}

@inproceedings{tian2016toward,
  title={Toward a severity index for ROP: An unsupervised approach},
  author={Tian, Peng and Ataer-Cansizoglu, Esra and Kalpathy-Cramer, Jayashree and Ostmo, Susan and Jonas, Karyn and Chan, RV Paul and Campbell, J Peter and Chiang, Michael F and Erdogmus, Deniz},
  booktitle={2016 38th Annual International Conference of the IEEE Engineering in Medicine and Biology Society (EMBC)},
  pages={1312--1315},
  year={2016},
  organization={IEEE}
}

@inproceedings{chen2018learning,
  title={Learning to explain: An information-theoretic perspective on model interpretation},
  author={Chen, Jianbo and Song, Le and Wainwright, Martin and Jordan, Michael},
  booktitle={International conference on machine learning},
  pages={883--892},
  year={2018},
  organization={PMLR}
}

@inproceedings{yoon2018invase,
  title={INVASE: Instance-wise variable selection using neural networks},
  author={Yoon, Jinsung and Jordon, James and van der Schaar, Mihaela},
  booktitle={International Conference on Learning Representations},
  year={2018}
}

@inproceedings{duchi2008efficient,
  title={Efficient projections onto the l 1-ball for learning in high dimensions},
  author={Duchi, John and Shalev-Shwartz, Shai and Singer, Yoram and Chandra, Tushar},
  booktitle={Proceedings of the 25th international conference on Machine learning},
  pages={272--279},
  year={2008}
}

@article{takapoui2020simple,
  title={A simple effective heuristic for embedded mixed-integer quadratic programming},
  author={Takapoui, Reza and Moehle, Nicholas and Boyd, Stephen and Bemporad, Alberto},
  journal={International journal of control},
  volume={93},
  number={1},
  pages={2--12},
  year={2020},
  publisher={Taylor \& Francis}
}

@inproceedings{leng2018extremely,
  title={Extremely low bit neural network: Squeeze the last bit out with admm},
  author={Leng, Cong and Dou, Zesheng and Li, Hao and Zhu, Shenghuo and Jin, Rong},
  booktitle={Proceedings of the AAAI conference on artificial intelligence},
  volume={32},
  number={1},
  year={2018}
}

@inproceedings{li2019admm,
  title={ADMM-based weight pruning for real-time deep learning acceleration on mobile devices},
  author={Li, Hongjia and Liu, Ning and Ma, Xiaolong and Lin, Sheng and Ye, Shaokai and Zhang, Tianyun and Lin, Xue and Xu, Wenyao and Wang, Yanzhi},
  booktitle={Proceedings of the 2019 on Great Lakes Symposium on VLSI},
  pages={501--506},
  year={2019}
}

@article{xu2018structured,
  title={Structured adversarial attack: Towards general implementation and better interpretability},
  author={Xu, Kaidi and Liu, Sijia and Zhao, Pu and Chen, Pin-Yu and Zhang, Huan and Fan, Quanfu and Erdogmus, Deniz and Wang, Yanzhi and Lin, Xue},
  journal={arXiv preprint arXiv:1808.01664},
  year={2018}
}

@inproceedings{yildiz2020fast,
  title={Fast and accurate ranking regression},
  author={Yildiz, Ilkay and Dy, Jennifer and Erdogmus, Deniz and Kalpathy-Cramer, Jayashree and Ostmo, Susan and Campbell, J Peter and Chiang, Michael F and Ioannidis, Stratis},
  booktitle={International Conference on Artificial Intelligence and Statistics},
  pages={77--88},
  year={2020},
  organization={PMLR}
}

@inproceedings{10.1117/12.2511964,
author = {Hristina Uzunova and Jan Ehrhardt and Timo Kepp and Heinz Handels},
title = {{Interpretable explanations of black box classifiers applied on medical images by meaningful perturbations using variational autoencoders}},
volume = {10949},
booktitle = {Medical Imaging 2019: Image Processing},
editor = {Elsa D. Angelini and Bennett A. Landman},
organization = {International Society for Optics and Photonics},
publisher = {SPIE},
pages = {1094911},
year = {2019},
doi = {10.1117/12.2511964},
URL = {https://doi.org/10.1117/12.2511964}
}

@inproceedings{lenis2020domain,
  title={Domain aware medical image classifier interpretation by counterfactual impact analysis},
  author={Lenis, Dimitrios and Major, David and Wimmer, Maria and Berg, Astrid and Sluiter, Gert and B{\"u}hler, Katja},
  booktitle={Medical Image Computing and Computer Assisted Intervention--MICCAI 2020: 23rd International Conference, Lima, Peru, October 4--8, 2020, Proceedings, Part I 23},
  pages={315--325},
  year={2020},
  organization={Springer}
}

@article{lecun1998mnist,
  title={The MNIST database of handwritten digits},
  author={LeCun, Yann},
  journal={http://yann. lecun. com/exdb/mnist/},
  year={1998}
}

@article{lecun2015lenet,
  title={LeNet-5, convolutional neural networks},
  author={LeCun, Yann and others},
  journal={URL: http://yann. lecun. com/exdb/lenet},
  volume={20},
  number={5},
  pages={14},
  year={2015}
}

@article{xiao2017fashion,
  title={Fashion-mnist: a novel image dataset for benchmarking machine learning algorithms},
  author={Xiao, Han and Rasul, Kashif and Vollgraf, Roland},
  journal={arXiv preprint arXiv:1708.07747},
  year={2017}
}

@article{adebayo2018sanity,
  title={Sanity checks for saliency maps},
  author={Adebayo, Julius and Gilmer, Justin and Muelly, Michael and Goodfellow, Ian and Hardt, Moritz and Kim, Been},
  journal={Advances in neural information processing systems},
  volume={31},
  year={2018}
}

@article{paszke2019pytorch,
  title={Pytorch: An imperative style, high-performance deep learning library},
  author={Paszke, Adam and Gross, Sam and Massa, Francisco and Lerer, Adam and Bradbury, James and Chanan, Gregory and Killeen, Trevor and Lin, Zeming and Gimelshein, Natalia and Antiga, Luca and others},
  journal={Advances in neural information processing systems},
  volume={32},
  year={2019}
}

@inproceedings{tian2019severity,
  title={A severity score for retinopathy of prematurity},
  author={Tian, Peng and Guo, Yuan and Kalpathy-Cramer, Jayashree and Ostmo, Susan and Campbell, John Peter and Chiang, Michael F and Dy, Jennifer and Erdogmus, Deniz and Ioannidis, Stratis},
  booktitle={Proceedings of the 25th ACM SIGKDD International Conference on Knowledge Discovery \& Data Mining},
  pages={1809--1819},
  year={2019}
}

@inproceedings{sundararajan2017axiomatic,
  title={Axiomatic attribution for deep networks},
  author={Sundararajan, Mukund and Taly, Ankur and Yan, Qiqi},
  booktitle={International conference on machine learning},
  pages={3319--3328},
  year={2017},
  organization={PMLR}
}

@inproceedings{shrikumar2017learning,
  title={Learning important features through propagating activation differences},
  author={Shrikumar, Avanti and Greenside, Peyton and Kundaje, Anshul},
  booktitle={International conference on machine learning},
  pages={3145--3153},
  year={2017},
  organization={PMlR}
}

@article{abnar2020quantifying,
  title={Quantifying attention flow in transformers},
  author={Abnar, Samira and Zuidema, Willem},
  journal={arXiv preprint arXiv:2005.00928},
  year={2020}
}

@inproceedings{chefer2021transformer,
  title={Transformer interpretability beyond attention visualization},
  author={Chefer, Hila and Gur, Shir and Wolf, Lior},
  booktitle={Proceedings of the IEEE/CVF conference on computer vision and pattern recognition},
  pages={782--791},
  year={2021}
}

@inproceedings{chefer2021generic,
  title={Generic attention-model explainability for interpreting bi-modal and encoder-decoder transformers},
  author={Chefer, Hila and Gur, Shir and Wolf, Lior},
  booktitle={Proceedings of the IEEE/CVF international conference on computer vision},
  pages={397--406},
  year={2021}
}

@inproceedings{belharbi2022f,
  title={F-cam: Full resolution class activation maps via guided parametric upscaling},
  author={Belharbi, Soufiane and Sarraf, Aydin and Pedersoli, Marco and Ben Ayed, Ismail and McCaffrey, Luke and Granger, Eric},
  booktitle={Proceedings of the IEEE/CVF Winter Conference on Applications of Computer Vision},
  pages={3490--3499},
  year={2022}
}
}

\appendix
\clearpage

\begin{center}
    \LARGE \textbf{Supplementary Material for} \\[1ex]
     \LARGE \textbf{``SSplain: Sparse and Smooth Explainer for Retinopathy of Prematurity
Classification''} \\
\end{center}
\section{Evaluation metrics}
\label{sec:supp_evaluation metrics}

We define three different metrics to evaluate SSplain on the ROP dataset, in addition to the post-hoc balance accuracy metric.  ``Connected Components'' evaluates smoothness. The ``Curvature`` and ``Dilation'' metrics assess whether the explainer detects tortuous and dilated regions, which are features that clinicians find important~\cite{chiang2021international}.

\subsection{Connected components}
\label{sec:supp_connected_components}
With this metric, we aim to measure the smoothness of explanation maps, focusing particularly on the smoothness of vessel structures at each insertion step $s$. To achieve this, we extract graphs from the image $\mathbf{X}$ and from images during the insertion process $\mathbf{X}_s$, and identify the length of the largest connected component in each graph. This length is the number of nodes in the largest connected component. 

We generate undirected graphs from images by extracting horizontal and vertical edges based on adjacent pixel values. Specifically, horizontal edge pairs $h_{edges}$ and vertical edge pairs $v_{edges}$, which define the start and end positions of edges, are given by:

\begin{equation}
\begin{aligned}
    h_{edges} &= \{[(j,k),(j+1,k)]| X_{j,k} \neq X_{j+1,k}\}, \\
    v_{edges} &= \{[(j,k),(j,k+1)]| X_{j,k} \neq X_{j,k+1}\}.
\end{aligned}
\end{equation}
 We create the graph using the NetworkX library~\cite{hagberg2008exploring} from these edge pairs. To identify the connected components, i.e., the sets of nodes that are connected to each other in the graph, we use the ``connected\_components'' function from NetworkX. This gives us a list of all connected components. We then calculate the length of the largest connected component, which is the number of nodes in that component. 
 
 Finally, we compute the ratio between the length of the largest connected components from $\mathbf{X}_s$ and $\mathbf{X}$. A higher value of the largest connected component at $s$ indicates that the inserted pixels create smooth vessel structures.

\subsection{Curvature}
\label{sec:supp_curvature}
Due to the link between retinal vessel tortuosity and ROP development, we evaluate the explainer’s ability to capture tortuous regions at each insertion step $s$. We compute pixel-wise curvatures $\mathbf{C}\in \reals_+^{h \times w}$ for $\mathbf{X}$ using the Tubular Curvature Filter Method~\cite{sunger2023tubular}, where higher pixel-wise curvature values indicate greater tortuosity. Sunger \etal~\cite{sunger2023tubular} calculates pixel-wise curvature values by measuring the directional rate of change in the eigenvectors. For more details on the curvature calculation, please refer to the work by Sunger \etal~\cite{sunger2023tubular}. 

Then, $\text{cos}(\mathbf{X}_s,\mathbf{C}) = \frac{\mathbf{X}_s \cdot \mathbf{C}}{\rVert\mathbf{X}_s\lVert \rVert\mathbf{C}\lVert}$ computes the similarity.
A higher value at $s$ shows that the inserted pixels capture tortuous regions, and doing this at an earlier $s$ indicates that the explainer assigns greater importance to these regions.

\subsection{Dilation}
\label{sec:supp_dilation}
We utilize the Average Segment Diameter (ASD) feature proposed by Ataer \etal~\cite{ataer2015computer} to assess how well the explainer captures dilation features at each insertion step $s$. ASD computes the total number of pixels in a vessel branch divided by the curve length. 
Specifically, 
\begin{equation}
    \text{ASD}(x_{\text{branch}}) = \frac{N(x_{\text{branch}})}{L_c(x_{\text{branch}})}\, ,
\end{equation}
where $N(x_{\text{branch}})$ is the pixel count on the vessel segment $x_{\text{branch}}$, and 
$L_c(x_{\text{branch}})$ is the length of $x_{\text{branch}}$ given in Eq.~\eqref{eq:curve_length}.
\begin{equation}
\label{eq:curve_length}
    L_c(x_{\text{branch}}) = \int_{a}^{b} \|\mathbf{c}'(t)\| \, \mathrm{d}t \,,
\end{equation}
parameterized by $\mathbf{c}(t): [a, b] \subset \mathbb{R} \mapsto \mathbb{R}^{2}$. We follow the tracing process defined by Ataer \etal~\cite{ataer2015computer} and  Tian \etal~\cite{tian2019severity} to extract vessel branches $x_{\text{branch}}$. Their tracing algorithm consists of two steps. First, they find sample points on vessel center-lines using a principal curve-based method. Then, they apply a minimum spanning tree algorithm to obtain tree structures and extract vessel branches.

Finally, we evaluate $\text{cos}(\mathbf{D_s},\mathbf{D})$, where $\mathbf{D}\in \mathbb{R}^{b\times 1}$ contains the ASD features from the original image $\mathbf{X}$ and $\mathbf{D}_s$ contains the features from the images during the insertion process  $\mathbf{X}_s$. Here, $b$ is the number of vessel branches. 

We extract $x_{\text{branch}}$ from the original image $\mathbf{X}$ to calculate both $\mathbf{D}$ and $\mathbf{D}_s$,  so only the number of pixels on each branch changes with changing $\mathbf{X}_s$. Similar to the curvature metric, a higher value at an earlier $s$ indicates that the explainer assigns greater importance to dilated vessel regions.

\begin{figure*}[t!]
    \centering    
    \centering
    \includegraphics[width=1.0\linewidth]{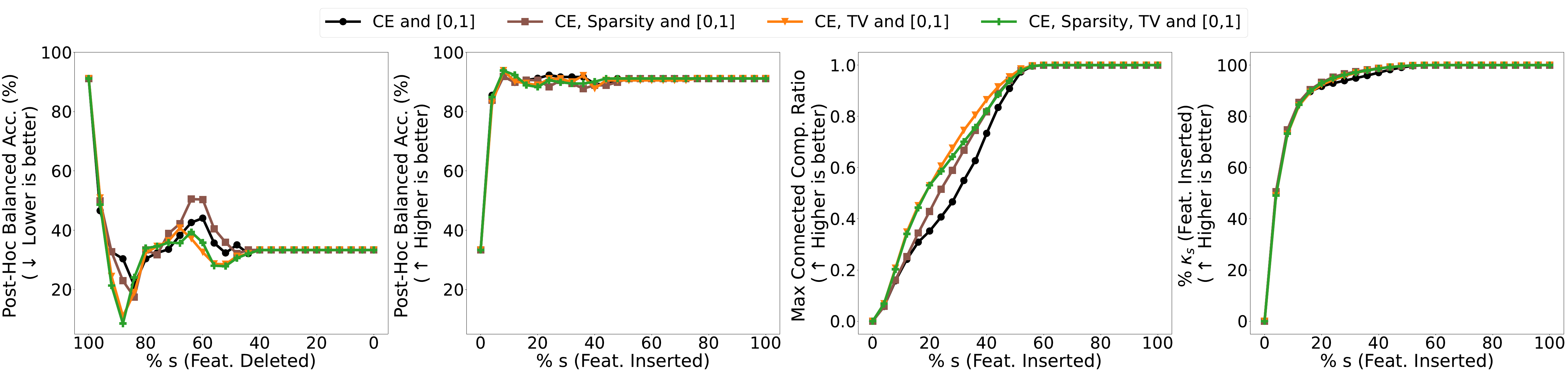}
    \caption{Ablation study for sparsity (using $\ell_0$ constraint) and total variation (tv) constraints on the ROP dataset. From left to right, we report the average of: Post-hoc balanced accuracy with deletion (lower is better) and insertion (higher is better) of pixels with the highest attribution scores, connected components ratio during the insertion process, and sparsity $s$ versus normalized sparsity $\kappa_s$ during the insertion process. Note that $S_0$ and $S_2$ constraints are always applied. Using all components in the problem formulation: cross-entropy loss, sparsity constrain $S_1$, total variation loss, and the $[0,1]$ ($S_2$) constraint together improves performance on all metrics.}
    \label{fig:ablation study}
\end{figure*}
\begin{figure*}[t!]
    \centering    
\includegraphics[width=1.0\linewidth]{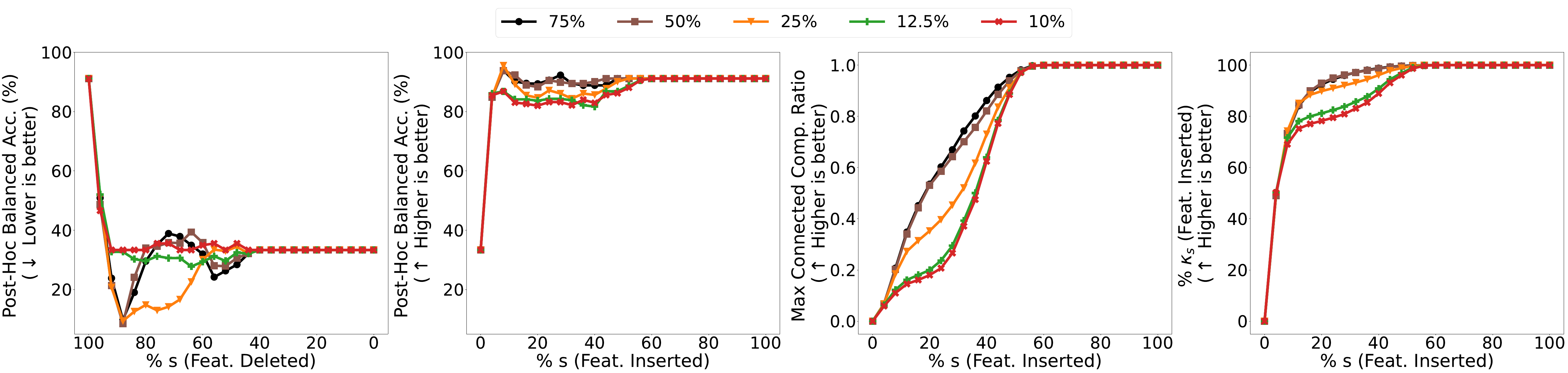}
    \caption{Comparison of different $\alpha$ for SSplain ($S_1$ with $\ell_0$ constraint): $75\%, 50\%, 25\%, 12.5\%$ and $10\%$ of $\lVert \mathbf{X} \rVert _{0}$ on the ROP dataset. From left to right, we report the average of: Post-hoc balanced accuracy with deletion (lower is better) and insertion (higher is better) of pixels with the highest attribution scores, connected components ratio during the insertion process, and sparsity $s$ versus normalized sparsity $\kappa_s$ during the insertion process. Making the sparsity constraint stricter ($\alpha$ smaller than $50\%$ of $\lVert \mathbf{X} \rVert _{0}$) results in performance loss.}
    \label{fig:sparsity_main}
\end{figure*}

\section{Ablation study}
\label{sec:ablation study}
The problem formulation in Eq.\eqref{eq:masking} from Section~\ref{sec:problem_formulation} includes a loss function $l(\cdot)$ (we use the cross-entropy function), a total variation function $\nu(\cdot)$, and $S_1$ and $S_2$ constraints with the support $S_0$ on pixels of vessels. Here, we present an ablation study on the ROP dataset to investigate the effects of sparsity, using $S_1$ with the $\ell_0$ norm and the total variation function, while always including the cross-entropy loss and the $S_2$ constraint (having mask values within $[0,1]$). Figure~\ref{fig:ablation study} shows that using this formulation improves performance, resulting in lower deletion accuracy, higher insertion accuracy and higher max connected component ratio.

\section{Sparsity area limit $\alpha$ for SSplain}
\label{sec:sparsity alpha}
In Eq.\eqref{eq:masking}, $S_1$ enforces the sparsity constraint, which is implemented using either the $\ell_0$ or $\ell_1$ norm. In Figure~\ref{fig:sparsity_main}, we use SSplain with the $\ell_0$ norm on the ROP dataset to compare different sparsity levels (upper bounds for the number of non-zero mask values): $75\%, 50\%, 25\%, 12.5\%$ and $10\%$ of $\lVert \mathbf{X} \rVert _{0}$.  We show that making this constraint stricter (by decreasing $\alpha$ in $S_1$) results in performance loss, i.e., higher deletion accuracy, lower insertion accuracy, a lower rate of connected components, and lower $\kappa_s$.

\section{Competing method details}
\label{sec:competing method details}
We compare SSplain with seven state-of-the-art explainers using the following settings in both our post-hoc performance analyses and visualizations on the ROP dataset:
\begin{itemize}
    \item \textbf{Gradient-based:}
    \begin{enumerate}
        \item \textit{Saliency~\cite{simonyan2013deep}:} The Saliency method is defined as the absolute value of the partial derivative of the $y$-th output neuron with respect to the input image, where $y$ is the target label.
        \item \textit{Guided Grad-CAM~\cite{selvaraju2017grad}:} It calculates the positive gradients of the target class with respect to the feature maps of a specific convolutional layer. One can apply the Guided Grad-CAM method to any layer. We use the output of the last Inception layer (``inception 5b'').
        \item \textit{Integrated Gradients~\cite{sundararajan2017axiomatic}:} It calculates the integral of the gradients along the path from a baseline input, such as a black image, to the input image.
        \item \textit{DeepSHAP~\cite{lundberg2017unified}:} DeepSHAP uses Shapley values from parts of the model to compute Shapley values for the whole model by propagating DeepLIFT~\cite{shrikumar2017learning}'s multipliers.
    \end{enumerate}
    \item \textbf{Perturbation-based:}
    \begin{enumerate}

        \item \textit{KernelSHAP~\cite{lundberg2017unified}:} It uses the LIME method to calculate Shapley values, which evaluates each feature's or input pixel's contribution to the prediction by considering all possible combinations, i.e., analyzes how the prediction changes when features are added to combinations. We use 200 perturbation samples.
        \item \textit{LIME~\cite{ribeiro2016should}:} It perturbs the image and trains a simple linear model with the given target label $t$ to learn feature importances using the perturbed images and their corresponding model predictions. We use 200 perturbation samples for this process.
        \item \textit{Occlusion ~\cite{zeiler2014visualizing}:} The Occlusion method measures feature importance by analyzing changes in the model's predictions when applying binary masks to image regions. The method uses a patch-wise mask that slides over the image. We use a patch size of $16\times16$ with $4\times4$ strides.
    \end{enumerate}
    \item \textbf{Incorporating sparsity or smoothness or both:}
    \begin{enumerate}
        \item \textit{Input$\times$Gradient ~\cite{shrikumar2016not}:} It calculates attribution scores by element-wise multiplication of the input with the gradients (Saliency scores) with respect to the target label $y$.
        \item \textit{Extremal Perturbation ~\cite{fong2019understanding}:} It defines a ranking-based area loss and uses a Gaussian kernel to incorporate smoothness to learn explanation maps by perturbing the input image. This perturbation involves either blurring or replacing pixel values with a constant value. The loss function includes preservation-based, deletion-based, or hybrid loss functions. Also, this method needs to start from an initial mask, which is initialized with all values set to one. For the ROP dataset, we use a Gaussian blur as the perturbation method, a ``hybrid'' loss with $400$ iterations, $0.05$ smoothing factor, and a target area to $50\%$ of $\lVert \mathbf{X} \rVert _{0}$. 
    \end{enumerate}
\end{itemize}

Even though the ROP images are in grayscale, the model accepts images with three channels, and some methods extract attribution for each channel. For visualization, we calculate the maximum of the channel-wise values for Saliency, Guided Grad-CAM, Integrated Gradients, Input$\times$Gradient, LIME, DeepSHAP and KernelSHAP. For evaluating connected components, curvature, and dilation, we report the channel-wise mean of these methods.

We use two different libraries to apply the competing methods to the ROP dataset. We use the Captum~\cite{kokhlikyan2020captum} library for all of the competitor methods except the Extremal Perturbation method. For the Extremal Perturbation method, we use TorchRay~\cite{fong2019understanding}.

\begin{figure*}[t!]
    \centering
    \includegraphics[width=1.0\linewidth]{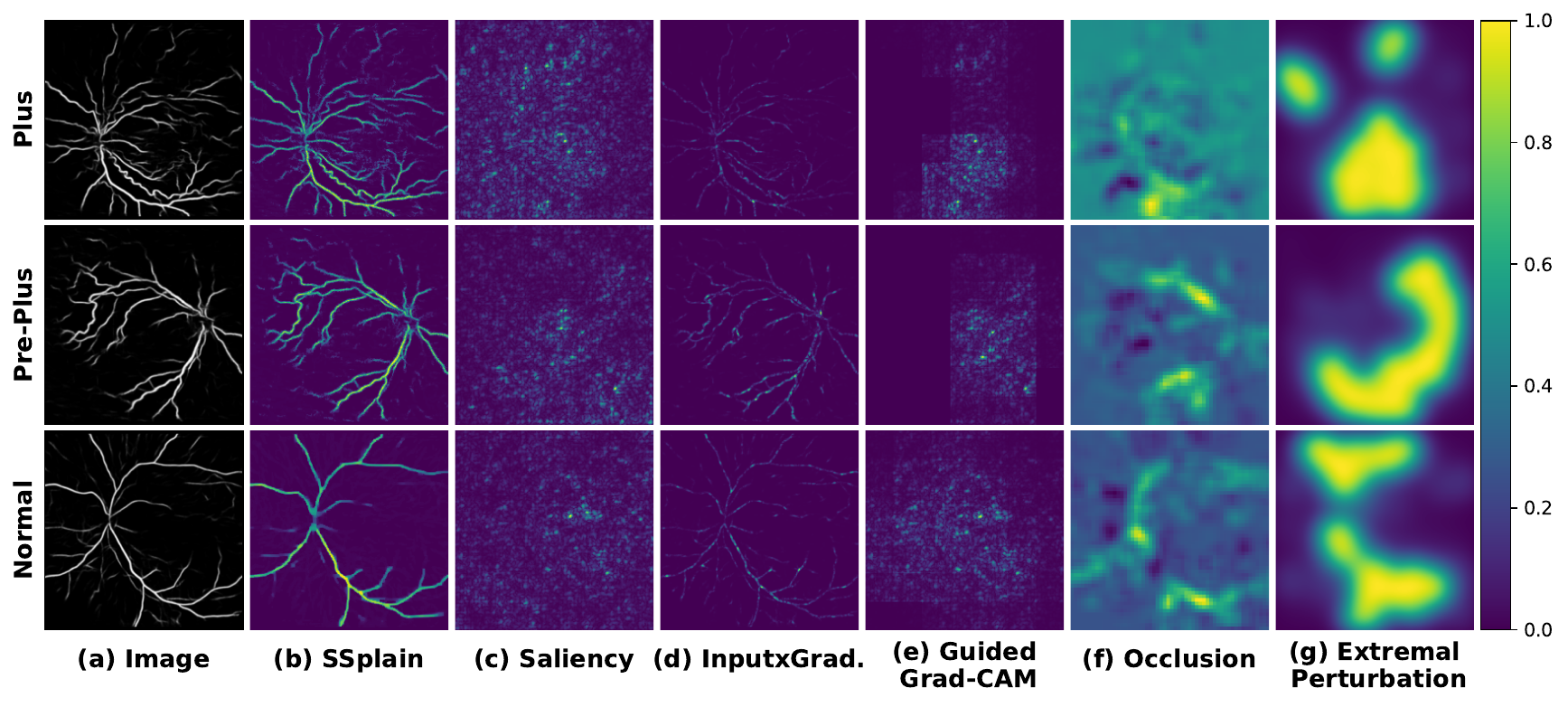}
    \caption{Comparison with other explainability methods using fundus images from the ROP dataset by visualizing the attribution scores. From left to right: (a) images, explanation maps generated using (b) SSplain ($S_1$ with $\ell_0$ constraint), (c) Saliency, (d) Input$\times$Gradient, (e) Guided Grad-CAM, (f) Occlusion and (g) Extremal Perturbation. We observe that SSplain effectively preserves image structures and assigns more importance to pixels with high tortuosity and dilation while maintaining smoothness. Except Input$\times$Gradient, other methods treat images as a whole and assign importance to pixels not associated with vessels.
    }
    \label{fig:methods_maps}
\end{figure*}

\begin{figure*}[t!]
    \centering
    \begin{subfigure}[b]{\textwidth}
    \includegraphics[width=1.0\linewidth]{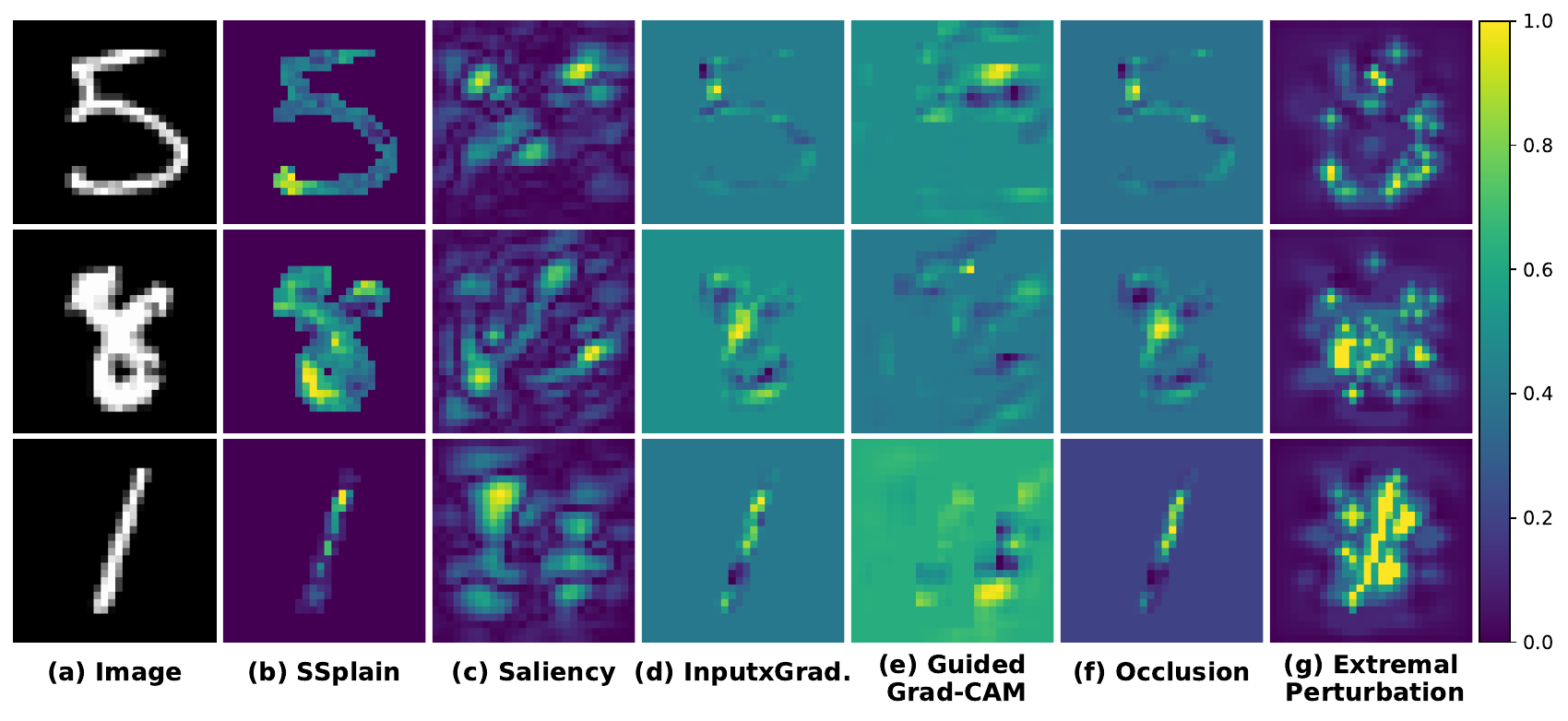}
    \caption{MNIST visualizations.}
    \end{subfigure}
    \begin{subfigure}[b]{\textwidth}
    \includegraphics[width=1.0\linewidth]{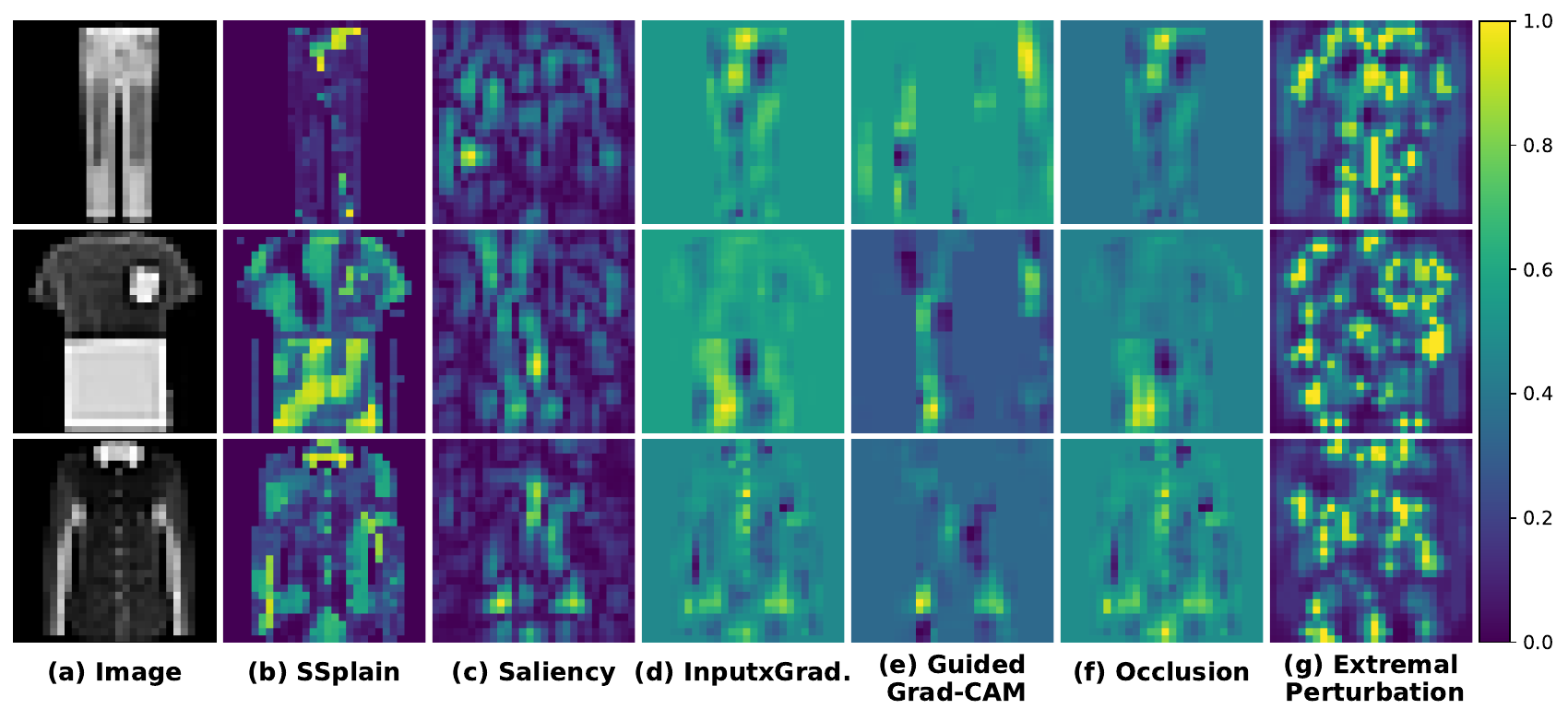}
    \caption{FMNIST visualizations.}
    \end{subfigure}
    \caption{Comparison of explainers on sample images from the MNIST and FMNIST datasets with attribution score visualization. From left to right: (a) images, explanation maps generated using (b) SSplain-0, (c) Saliency, (d) Input$\times$Gradient, (e) Guided Grad-CAM, (f) Occlusion and (g) Extremal Perturbation. 
    SSplain preserves sparse image structures while others assign importance to the background.
    }
    \label{fig:methods_maps_MNISTandFMNIST}
\end{figure*}

\section{Additional Dataset and Model Details}
\label{sec:additional_experiments}

In this section, we apply SSplain to the MNIST~\cite{lecun1998mnist} and FMNIST~\cite{xiao2017fashion} datasets. Similar to the ROP dataset, the background of the images consists of black pixels (zero-valued pixels), and we can define support $S_0$ as in Section~\ref{sec:problem_formulation} using the non-zero pixels. To obtain the trained black-box model for both of these datasets, we train the LeNet-5~\cite{lecun2015lenet} using the Adam optimizer~\cite{kingma2014adam} with a learning rate of 0.001, a weight decay of 0.0001, over 50 epochs, and a batch size of 32, with early stopping.

We compare SSplain with nine explainers. \textit{Gradient-based:} Saliency~\cite{simonyan2013deep}, Guided Grad-CAM~\cite{selvaraju2017grad}, Integrated Gradients~\cite{sundararajan2017axiomatic} and DeepSHAP~\cite{lundberg2017unified}. We apply Guided Grad-CAM to the last CNN layer.
\textit{Perturbation-based:}  LIME~\cite{ribeiro2016should}, KernelSHAP~\cite{lundberg2017unified} and Occlusion ~\cite{zeiler2014visualizing}. KernelSHAP and LIME employ $200$ input samples. Occlusion uses $1\times1$ windows.
\textit{Incorporating sparsity or smoothness or both:} Input$\times$Gradient ~\cite{shrikumar2016not} and Extremal Perturbation ~\cite{fong2019understanding}. Extremal Perturbation uses a Gaussian blur, a ``hybrid'' loss with $200$ iterations, $0.0001$ smoothing factor, and a target area to $25\%$ of $\lVert \mathbf{X} \rVert _{0}$ with having pixel-wise masks. It initializes mask values to be all ones.

To evaluate the performance of explainers, we use post-hoc accuracy for the insertion and deletion processes, and we also evaluate sparsity $s$ and normalized sparsity $\kappa_s$ during the insertion process. We do not use the domain-specific metrics for ROP, such as curvature and dilation.
\subsection{MNIST}
\label{sec:MNIST}
MNIST dataset~\cite{lecun1998mnist} has 70,000 $28 \times 28$ greyscale images (split into 60,000 for training and 10,000 for testing), each corresponding to one of the digits from 0 to 9. The trained model has $98.53\%$ accuracy on the test set. We perform our post-hoc explainability analysis on 500 sample images from the test set. For SSplain, we initialize masks with values proportional to the corresponding pixel intensity between 0 and 1, as follows: $\mathbf{M} = \frac{\mathbf{X}}{|\mathbf{X}|_{\infty}}$. We use the $\ell_0$ norm with $\alpha$ in $S_1$ set to $25\%$ of $\lVert \mathbf{X} \rVert _{0}$. We use the cross-entropy loss function for $l(\cdot,\cdot)$ and run 20 iterations of ADMM for each image. We utilize the Adam optimizer~\cite{kingma2014adam} to solve problem~\eqref{eq:updateM} with a learning rate of 0.1, $\rho=0.01$ and $\lambda=10^{-3}$.

\subsection{FMNIST}
FMNIST dataset~\cite{xiao2017fashion} consists of 70,000 $28 \times 28$ greyscale images (split into 60,000 for training and 10,000 for testing), each corresponding to fashion products from 10 different classes. The trained model achieves $88.67\%$ accuracy on the test set. We conduct our post-hoc explainability analysis on 500 sample images from the test set. For SSplain, we initialize masks with values proportional to the corresponding pixel intensity between 0 and 1, as follows: $\mathbf{M} = \frac{\mathbf{X}}{|\mathbf{X}|_{\infty}}$. We use the $\ell_0$ norm with $\alpha$ in $S_1$ set to $25\%$ of $\lVert \mathbf{X} \rVert _{0}$, the cross-entropy loss function for $l(\cdot,\cdot)$ and run 20 iterations of ADMM for each image. We use the Adam optimizer~\cite{kingma2014adam} to solve problem~\eqref{eq:updateM} with a learning rate of 0.1, $\rho=0.01$ and $\lambda=10^{-4}$.

\section{Visualization}
\label{sec:visualization}
\subsection{ROP}
Figure~\ref{fig:methods_maps} shows the explanation maps from SSplain (with $\ell_0$) and comparative methods (Saliency, Input$\times$Gradient, Guided Grad-CAM, Occlusion, and Extremal Perturbations) for sample images from the ROP dataset. We can see that SSplain gives more importance to tortuous and dilated vessel regions, while other methods fail to capture these regions. Except for Input$\times$Gradient, the other methods assign importance to background pixels because they treat the images as a whole.

\label{sec:supplemet visualization rop}

\subsection{MNIST and FMNIST}
\label{sec:vizualizationsforMNISTandFMNIST}
Figure~\ref{fig:methods_maps_MNISTandFMNIST} shows the explanation maps from SSplain (with $\ell_0$) and comparative methods (Saliency, Input$\times$Gradient, Guided Grad-CAM, Occlusion, and Extremal Perturbations) for sample images from the MNIST and FMNIST datasets. SSplain effectively preserves sparse image structures while other methods assign importance to background pixels.

\label{sec:supplemet visualization rop}

\section{Sparsity constraints ($\ell_0$ vs $\ell_1$)}
\label{sec:sparsity constraints}

\subsection{SSplain with $\ell_0$ and $\ell_1$ constraints}
\label{sec:S_1 with l0 and l1}
\begin{figure}[t!]
    \centering
    \includegraphics[width=1.0\linewidth]{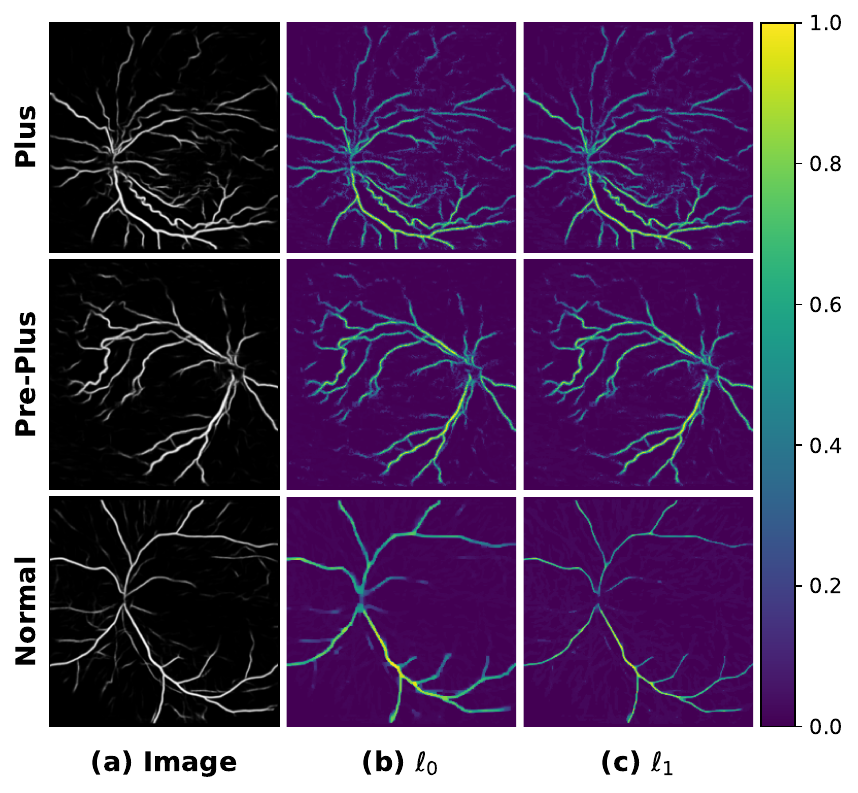}
    \caption{Comparison of the $\ell_0$ and $\ell_1$ norms in the sparsity constraint $S_1$ for the SSplain method, using fundus images from the ROP dataset to visualize the attribution scores. From left to right: (a) images, explanation maps generated using (b) SSplain with the $\ell_{0}$ constraint, (c) SSplain with the $\ell_{1}$ constraint. We initialize masks with values proportional to the corresponding pixel intensity between 0 and 1, as follows: $\mathbf{M} = \frac{\mathbf{X}}{|\mathbf{X}|_{\infty}}$. We run $50$ iterations with a learning rate of 0.01, $\rho=0.01$ and $\lambda=10^{-5}$ using the Adam optimizer. We set the sparsity level to $50\%$ of $\lVert \mathbf{X} \rVert _{0}$ for both cases. We do not observe any significant difference between these two variations in the $S_1$ constraint.}
    \label{fig:S1_l0_l1_visualization}
\end{figure}

SSplain can integrate both the $\ell_0$ and $\ell_1$ norms in its $S_1$ constraint. In Section~\ref{sec:experiments metrics analyses}, we showed that even though both variations work well, the $\ell_0$ variation performs slightly better for the deletion and insertion post-hoc accuracy analyses. Figure~\ref{fig:S1_l0_l1_visualization} displays the explanation maps for both SSplain-0 and SSplain-1 on the sample ROP images, showing no significant difference between them.

\subsection{SSplain vs Extremal Perturbations}
\label{sec:ssplain vs extremal perturbations}
\begin{figure}[h!]
    \centering
    \includegraphics[width=1.0\linewidth]{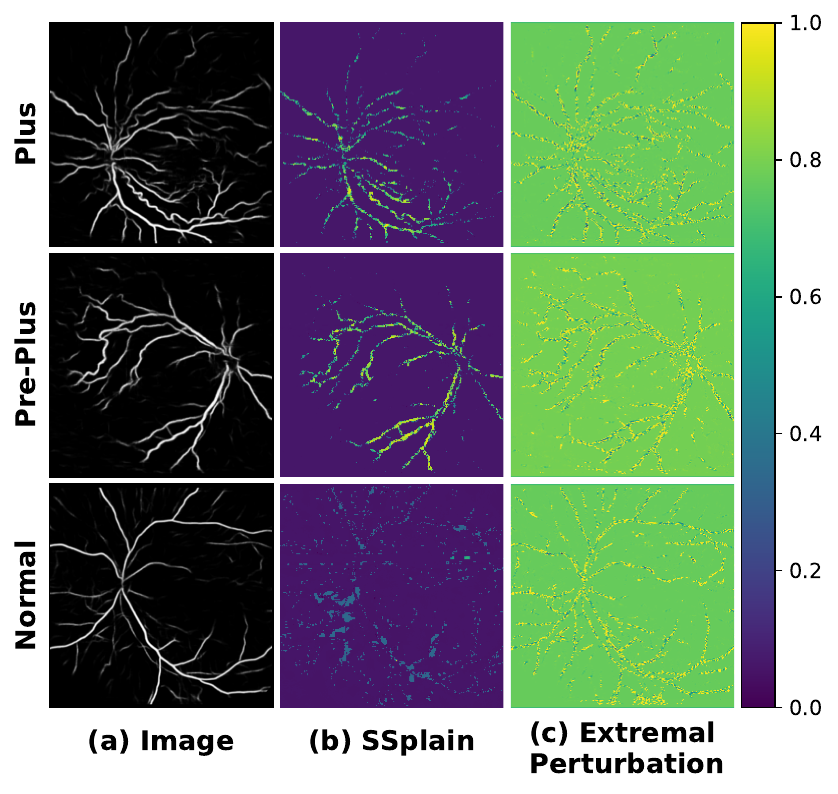}
    \caption{Comparison of SSplain-0 and Extremal Perturbation methods using fundus images from the ROP dataset by visualizing the attribution scores. From left to right: (a) images, explanation maps generated using (b) SSplain, Extremal Perturbation. Extremal Perturbation uses a ``preservation'' loss with  $0.0001$ smoothing factor, and apply pixel-wise masking. SSplain uses $\rho=0.01$, $\lambda = 0.0001$ and does not use the $S_0$ constraint. Both methods use $400$ iterations, a target area of $5\%$, a learning rate of $0.01$ and initialize mask values to be all ones. SSplain captures the support of vessels, tortuous and dilated regions. It does not assign importance to the background, even without the $S_0$ constraint, whereas Extremal Perturbation does.
    }
    \label{fig:sparsity_analyses}
\end{figure}

In this section, we compare SSplain with the $\ell_0$ norm constraint in $S_1$ and Extremal Perturbations methods, which use the $\ell_1$ constraint. Although SSplain has the $S_0$ constraint, for this experiment, we remove this constraint and assign initial mask valuesof all pixels to be ones in both methods to ensure a fair comparison. Both methods run for 400 iterations with a target area of $5\%$ and a learning rate of 0.01. SSplain uses $\rho=0.01$ and $\lambda=0.0001$. The Extremal Perturbation method uses a ``preservation'' loss with a 0.0001 smoothing factor and employs pixel-wise masks. It perturbs the image by replacing pixel values with a constant value of zero. Figure~\ref{fig:sparsity_analyses} shows that SSplain still creates explanations on the vessel pixels and does not assign importance to the background. It still gives more importance to tortuous and dilated vessel regions. However, the extremal perturbation method gives importance to the background, resulting in significantly less sparse masks.

\section{Execution Times}
\label{sec:execution time}

In Table~\ref{tab:execution time}, we report the execution times (mean $\pm$ standard deviation) for each dataset. We include only the execution times of input-perturbation-based approaches, which are KernelSHAP~\cite{lundberg2017unified}, LIME~\cite{ribeiro2016should}, Occlusion~\cite{zeiler2014visualizing}, and Extremal Perturbation~\cite{fong2019understanding}, since SSplain is part of that explainability method family. The gradient-based method requires only a single backpropagation on the model and is faster than input-perturbation-based explainability methods. For the ROP dataset, execution times are averaged over 100 images, while for MNIST and FMNIST, they are averaged over 500 images.  

Note that the runtime of all input perturbation methods depends on the configurations of the explainers and the black-box model size. SSplain’s runtime depends on the number of ADMM iterations, image size, and model size. Extremal Perturbations' runtime depends on the number of iterations, image size, patch size, and model size. KernelSHAP and LIME's runtime depends on the model size and the number of perturbation samples. Occlusion's runtime depends on the model size, image size, patch size, and stride. Please refer to Section~\ref{sec:competing_methods} for the configurations of the explainers for the ROP dataset and Appendix~\ref{sec:additional_experiments} for the MNIST and FMNIST datasets.

\begin{table}[t!]
\centering
\caption{Execution times (mean $\pm$ standard deviation) in terms of seconds for KernelSHAP~\cite{lundberg2017unified}, LIME~\cite{ribeiro2016should}, Occlusion~\cite{zeiler2014visualizing}, Extremal Perturbation~\cite{fong2019understanding}, and SSplain on the ROP, MNIST, and FMNIST datasets. For the ROP dataset, execution times are averaged over 100 images, while for MNIST and FMNIST, they are averaged over 500 images. We report only the execution times of input-perturbation-based approaches, since SSplain belongs to this family of methods. Gradient-based methods require only a single backpropagation through the model and are generally faster than input-perturbation-based approaches. Note that these execution times depend heavily on the explainer configuration, such as the number of iterations and patch size, when applicable.}
\label{tab:execution time}
\resizebox{1.0\columnwidth}{!}{\begin{tabular}{|l|l|l|l|}
\hline
\textbf{Methods}               & \textbf{ROP}       & \textbf{MNIST}  & \textbf{FMNIST} \\ \hline
\textit{KernelSHAP}            & 29.57 $\pm$ 7.24   & 0.23 $\pm$ 0.07 & 0.24 $\pm$ 0.02 \\ \hline
\textit{LIME}                  & 27.86 $\pm$ 2.36   & 0.21 $\pm$ 0.06 & 0.23 $\pm$ 0.03 \\ \hline
\textit{Occlusion}             & 114.60 $\pm$ 13.37 & 0.64 $\pm$ 0.05 & 0.64 $\pm$ 0.02 \\ \hline
\textit{\begin{tabular}[c]{@{}l@{}}Extremal \\ Perturbations\end{tabular}} & 222.36 $\pm$ 11.27 & 15.13 $\pm$ 0.39 & 15.19 $\pm$ 0.49 \\ \hline
\textit{SSplain}               & 29.91 $\pm$ 2.13   & 1.58 $\pm$ 0.04 & 1.59 $\pm$ 0.02  \\ \hline
\end{tabular}}
\end{table}

\section{Mask Initialization}
\label{sec:mask initialization}

\begin{figure}[t]
    \centering
    \includegraphics[width=1.0\linewidth]{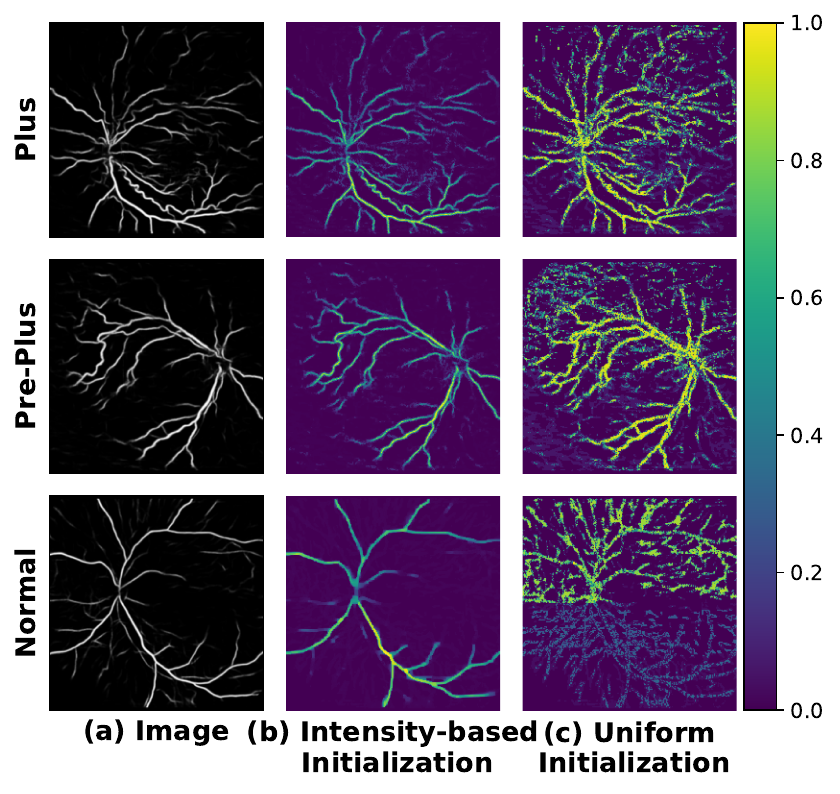}
    \caption{Comparison of SSplain-0 with intensity-based mask initialization versus uniform (all-ones) mask initialization using fundus images from the ROP dataset by visualizing the attribution scores. From left to right: (a) images, explanation maps generated using (b) SSplain-0 with intensity-based, and SSplain-0 with uniform mask initializations.  The intensity-based initialization case uses with $50$ ADMM iterations, while the uniform initialization uses $400$. Both methods use $\rho=0.01$, $\lambda = 0.00001$, a learning rate of $0.01$ and a target area of $50\%$ of $\alpha_0$ in $S_1$ to $50\%$ of $\lVert \mathbf{X} \rVert _{0}$. Uniform initialization results in less sparse and less smooth explanations compared to intensity-based initialization.
    }
    \label{fig:initialization_vis}
\end{figure}

\begin{figure*}[t!]
    \centering    
     \begin{subfigure}[b]{1.0\textwidth}
         \centering
        \includegraphics[width=1.0\linewidth]{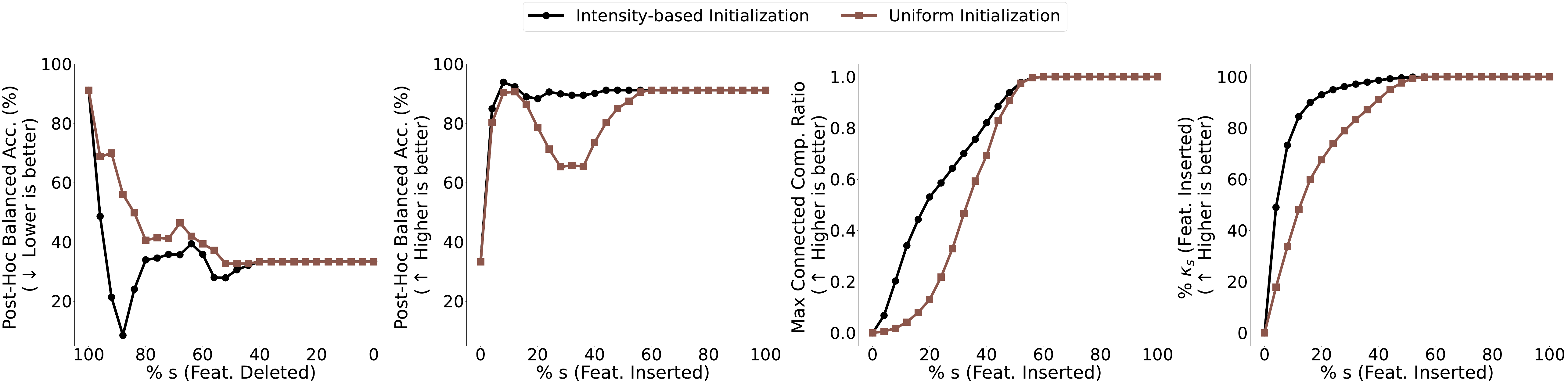}
         \caption{Accuracy, connected components, normalized sparsity $\kappa_s$ metrics vs.~sparsity $s$.}
         \label{fig:results_level_s}
     \end{subfigure}
      \begin{subfigure}[b]{1.0\textwidth}
         \centering         \includegraphics[width=1.0\linewidth]{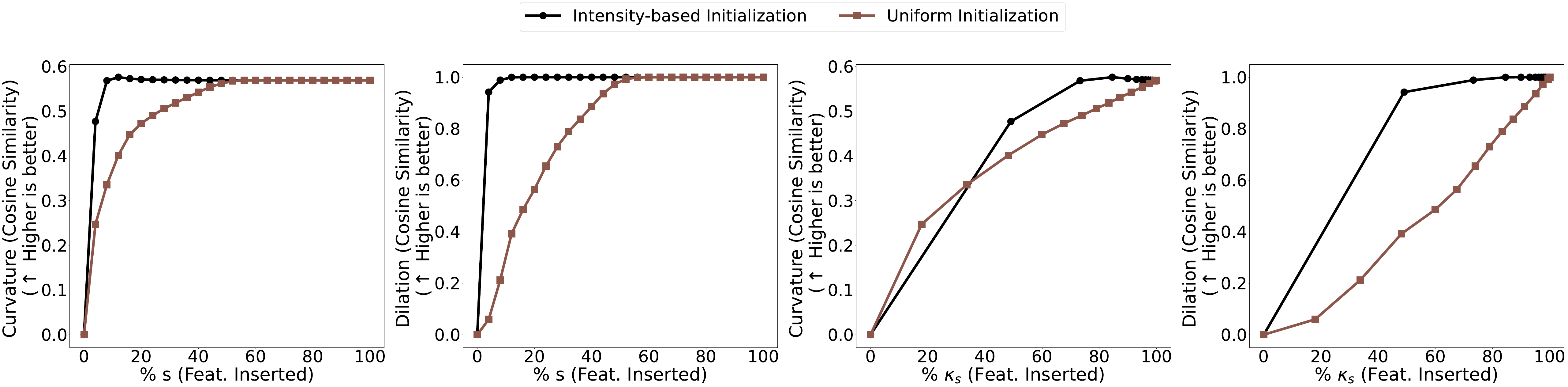}
         \caption{Curvature and dilation analyses vs.~sparsity $s$ and~normalized sparsity $\kappa_s$.}
         \label{fig:results_s_norm}
     \end{subfigure}
    \caption{Comparison SSplain-0 with intensity-based mask initialization versus uniform (all-ones) mask initialization for post-hoc analyses on the ROP dataset. The intensity-based case uses $50$ ADMM iterations, while the uniform case uses $400$. Both use $\rho=0.01$, $\lambda = 0.00001$, a learning rate of $0.01$ and a target area of $50\%$ of $\alpha_0$ in $S_1$ to $50\%$ of $\lVert \mathbf{X} \rVert _{0}$.(a) We report the average of: Post-hoc balanced accuracy with deletion (lower is better) and insertion (higher is better) of pixels with the highest attribution scores, connected components ratio with insertion, and sparsity $s$ versus normalized sparsity $\kappa_s$ during the insertion process. (b) Curvature and dilation similarity with the insertion process with respect to  $s$ and $\kappa_s$. Intensity-based mask initialization outperforms uniform mask initialization across all metrics.}
    \label{fig:initialization_posthoc}
\end{figure*}

We compare initializations for mask $\mathbf{M}$. In the main paper, we initialize masks with values proportional to the corresponding pixel intensity between 0 and 1, as follows: $\mathbf{M} = \frac{\mathbf{X}}{\|\mathbf{X}\|_{\infty}}$. The mask initialization depends on the use case, and for our purposes of optimizing a mask with sparsity and smoothness constraints, this intensity-based initialization is suitable. Alternatively, one can initialize masks as all ones assigned to the support of the image using $S_0$, where $M^0_{j,k} = 1$ for $M_{j,k} \in S_0$ with $S_0 = \{\mathbf{M} : \supp(M_{jk}) \subseteq \supp (X_{jk}) \}$. $\supp(\cdot)$ denotes the support (i.e., the non-zero pixels). We compare these two cases: intensity-based and uniform (all-ones) initialization on the ROP dataset.

Both methods use the $S_0$ constraint and the $\ell_0$ constraint for sparsity in $S_1$. The intensity-based initialization case uses with $50$ ADMM iterations, while the uniform initialization uses $400$. This iteration difference is since the uniform case starts from a less sparse point. Both methods use $\rho=0.01$, $\lambda = 0.00001$, a learning rate of $0.01$ and a target area of $50\%$ of $\alpha_0$ in $S_1$ to $50\%$ of $\lVert \mathbf{X} \rVert _{0}$. 

\subsection{Visualizations}
Here, we visually compare explainability maps from different mask initialization methods. Figure~\ref{fig:initialization_vis} shows that uniform initialization results in less sparse and less smooth explanations compared to intensity-based initialization on sample ROP images.

\subsection{Post-hoc Analyses}
Figure~\ref{fig:initialization_posthoc} shows a comparison of mask initialization on the ROP dataset. This evaluation uses post-hoc balanced accuracy, sparsity, number of connected components, and disease-related metrics of tortuosity and dilation. We observe that intensity-based mask initialization outperforms uniform mask initialization across all metrics.

\section{Unsupervised Generation}
\label{sec:unsupervised_generation}
Here, we follow an unsupervised approach where the target class $y$ for generating explanations in Eq.~\ref{eq:masking} is the model's predicted class. Formally, $y = \arg\max_{c} f_c(\mathbf{X})$, where $f:\mathbb{R}^{h\times w}\to \mathbb{R}^c,\mathbf{X}\mapsto f(\mathbf{X})$ is the black-box model, with $c$ the number of classes and $\mathbf{X} \in \mathbb{R}^{h \times w}$ the input image, with $h$ and $w$ denote the image height and width.

This section includes unsupervised explanation generations and their evaluation using accuracy-based metrics. SSplain and competing explainer setups are given in Sections~\ref{sec:datasetandmodel} and~\ref{sec:competing_methods} for the ROP dataset, and in Appendix~\ref{sec:additional_experiments} for the MNIST and FMNIST datasets.

\begin{figure*}[t]
    \centering    
     \centering
     \includegraphics[width=1.0\linewidth]{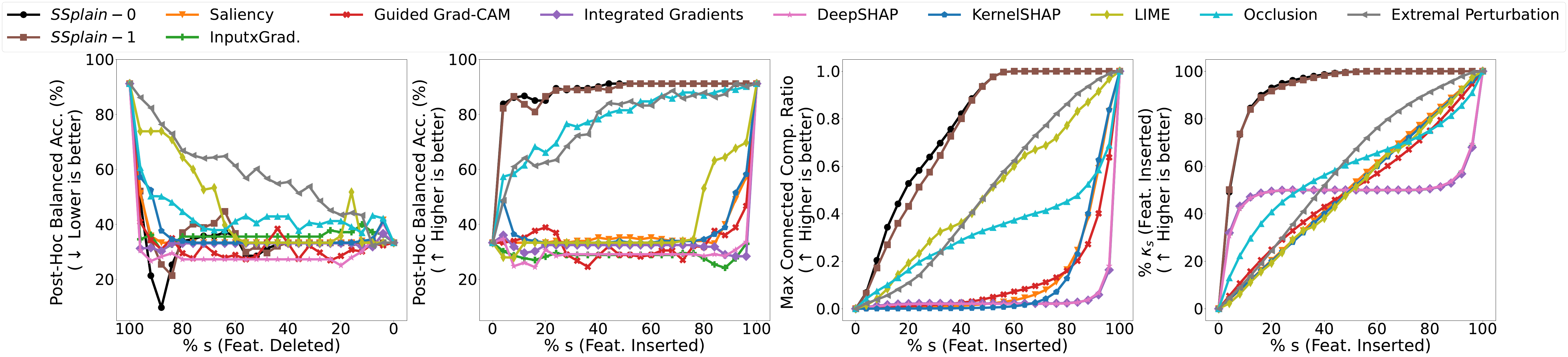}
    \caption{Comparison of explainers on the ROP dataset for unsupervised explaination generation: SSplain-0 ($S_1$ with $\ell_0$ constraint), SSplain-1 ($S_1$ with $\ell_1$ constraint), Saliency~\cite{simonyan2013deep}, Input$\times$Gradient~\cite{shrikumar2016not}, Guided Grad-CAM~\cite{selvaraju2017grad},  Integrated Gradients~\cite{sundararajan2017axiomatic}, DeepSHAP~\cite{lundberg2017unified}, KernelSHAP~\cite{lundberg2017unified}, LIME~\cite{ribeiro2016should}, Occlusion~\cite{zeiler2014visualizing} and Extremal Perturbation~\cite{fong2019understanding}.  We report the average of: Post-hoc balanced accuracy with deletion (lower is better) and insertion (higher is better) of pixels with the highest attribution scores, connected components ratio during the insertion process, and sparsity $s$ versus normalized sparsity $\kappa_s$ during the insertion process.  Similar to the supervised explanation generation, SSplain consistently outperforms competitors and competitors give importance to the background. }
    \label{fig:unsupervised_ROP}
\end{figure*}

\begin{figure*}[t]
    \centering    
     \centering
     \includegraphics[width=0.75\linewidth]{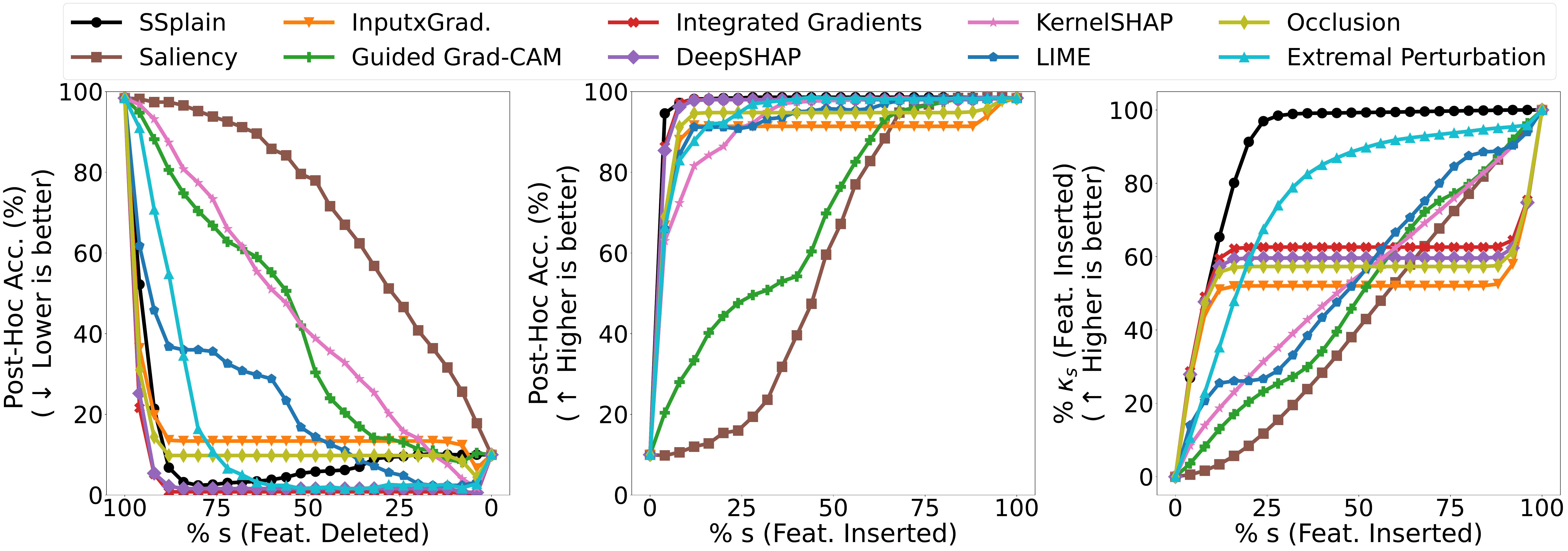}
    \caption{Comparison of explainers on the MNIST dataset for unsupervised explaination generation: SSplain-0 ($S_1$ with $\ell_0$ constraint), SSplain-1 ($S_1$ with $\ell_1$ constraint), Saliency~\cite{simonyan2013deep}, Input$\times$Gradient~\cite{shrikumar2016not}, Guided Grad-CAM~\cite{selvaraju2017grad},  Integrated Gradients~\cite{sundararajan2017axiomatic}, DeepSHAP~\cite{lundberg2017unified}, KernelSHAP~\cite{lundberg2017unified}, LIME~\cite{ribeiro2016should}, Occlusion~\cite{zeiler2014visualizing} and Extremal Perturbation~\cite{fong2019understanding}.  We report the average of: Post-hoc accuracy with deletion (lower is better) and insertion (higher is better) of pixels with the highest attribution scores, connected components ratio during the insertion process, and sparsity $s$ versus normalized sparsity $\kappa_s$ during the insertion process. Consistent with the supervised generation, SSplain-0 consistently outperforms the competing methods in all analyses, achieving lower deletion and higher insertion post-hoc accuracy, along with higher normalized sparsity $\kappa_s$. }
    \label{fig:unsupervised_MNIST}
\end{figure*}

\begin{figure*}[t!]
    \centering    
     \centering
     \includegraphics[width=0.75\linewidth]{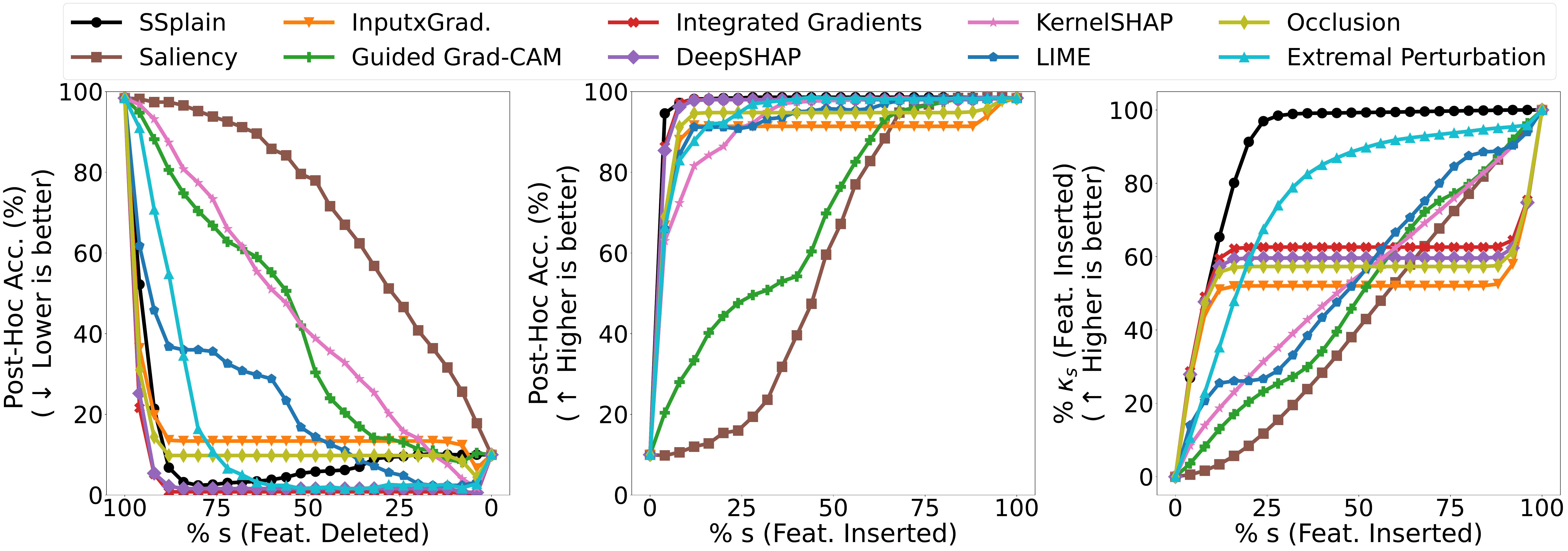}
    \caption{Comparison of explainers on the FMNIST dataset for unsupervised explaination generation: SSplain-0 ($S_1$ with $\ell_0$ constraint), SSplain-1 ($S_1$ with $\ell_1$ constraint), Saliency~\cite{simonyan2013deep}, Input$\times$Gradient~\cite{shrikumar2016not}, Guided Grad-CAM~\cite{selvaraju2017grad},  Integrated Gradients~\cite{sundararajan2017axiomatic}, DeepSHAP~\cite{lundberg2017unified}, KernelSHAP~\cite{lundberg2017unified}, LIME~\cite{ribeiro2016should}, Occlusion~\cite{zeiler2014visualizing} and Extremal Perturbation~\cite{fong2019understanding}.  We report the average of: Post-hoc accuracy with deletion (lower is better) and insertion (higher is better) of pixels with the highest attribution scores, connected components ratio during the insertion process, and sparsity $s$ versus normalized sparsity $\kappa_s$ during the insertion process. Consistent with the supervised generation, although SSplain-0 excels in insertion-based accuracy and normalized sparsity $\kappa_s$ analyses, Occlusion performs better than SSplain in deletion analysis. However, SSplain still stabilizes earlier than all competing methods.}
    \label{fig:unsupervised_FMNIST}
\end{figure*}
\subsection{Post-hoc Analyses}

\subsubsection{ROP Dataset}
Figure~\ref{fig:unsupervised_ROP} shows unsupervised explanation generation on the ROP dataset. Consistent with the supervised setup in Section~\ref{sec:posthoc}, SSplain outperforms the competitors by achieving lower deletion, higher insertion accuracy, a higher connected components ratio, and higher normalized sparsity $\kappa_s$, while also stabilizing earlier than the other methods.

\subsubsection{MNIST Dataset}
Figure~\ref{fig:unsupervised_MNIST} demonstrates unsupervised explanation generation across different analyses and metrics for the MNIST dataset. Similar to the supervised generation, SSplain-0 consistently outperforms the competing methods in all analyses, achieving lower deletion and higher insertion post-hoc accuracy, along with higher normalized sparsity $\kappa_s$. Moreover, SSplain reaches its maximum earlier than other methods and remains stable throughout the insertion process.

\subsubsection{FMNIST Dataset}

Figure~\ref{fig:unsupervised_FMNIST} presents the performance of explainers in an unsupervised setup across different analyses and metrics for the FMNIST dataset. Consistent with the supervised setup in Section~\ref{sec:add_posthoc}, although SSplain-0 excels in insertion-based accuracy and normalized sparsity $\kappa_s$ analyses, Occlusion performs better than SSplain in deletion analysis. However, SSplain still stabilizes earlier than all competing methods.

\end{document}